\documentclass[letterpaper,twocolumn,fleqn]{article}

\usepackage{packages/ist}
\usepackage{times}
\usepackage{wrapfig}
\usepackage{graphicx}
\usepackage{amsmath}
\usepackage{amssymb}
\usepackage{adjustbox}
\usepackage{booktabs}
\usepackage[style=base]{caption}
\usepackage{multirow}
\usepackage{multicol}
\usepackage{xspace}
\usepackage[table]{xcolor}

\makeatletter
\let\NAT@parse\undefined
\DeclareRobustCommand\onedot{\futurelet\@let@token\@onedot}
\def\@onedot{\ifx\@let@token.\else.\null\fi\xspace}
 
\def\ie{\emph{i.e}\onedot} 
\def\etal{\emph{et al}\onedot}
\makeatother

\usepackage[pagebackref=true,breaklinks=true,letterpaper=true,colorlinks,bookmarks=false]{hyperref}
\title{Adversarial Attacks on Multi-task Visual Perception for Autonomous Driving}

\author{
Ibrahim Sobh$^{1}$,
Ahmed Hamed$^{1}$,
Varun Ravi Kumar$^{2, 3}$ and
Senthil Yogamani$^{4}$  \\
{$^{1}$Valeo R\&D Egypt \\
$^{2}$Valeo DAR Germany \\
$^{3}$Technische Universit\"at Ilmenau, Germany \\
$^{4}$Valeo Ireland}}

\begin{document}
\maketitle 
\thispagestyle{empty}
% -------------------------------------------------
\begin{abstract}
Deep neural networks (DNNs) have accomplished impressive success in various applications, including autonomous driving perception tasks, in recent years. On the other hand, current deep neural networks are easily fooled by adversarial attacks. This vulnerability raises significant concerns, particularly in safety-critical applications. As a result, research into attacking and defending DNNs has gained much coverage. In this work, detailed adversarial attacks are applied on a diverse multi-task visual perception deep network across distance estimation, semantic segmentation, motion detection, and object detection. The experiments consider both white and black box attacks for targeted and un-targeted cases, while attacking a task and inspecting the effect on all the others, in addition to inspecting the effect of applying a simple defense method. We conclude this paper by comparing and discussing the experimental results, proposing insights and future work. The visualizations of the attacks are available at {\url{https://youtu.be/6AixN90budY}}.
\end{abstract}
% -------------------------------------------------
% -------------------------------------------------
\section{Introduction}

Autonomous Vehicles are expected to significantly reduce accidents~\cite{avsafetywho}, where visual perception systems~\cite{dahal2021roadedgenet, dahal2021online, das2020tiledsoilingnet, dhananjaya2021weather, gallagher2021hybrid, uricar2021let, yahiaoui2019fisheyemodnet} are in the heart of these vehicles. Despite the notable achievements of DNNs in visual perception, we can easily fool the networks by adversarial examples that are imperceptible to the human eye but cause the network to fail. Adversarial examples are usually created by deliberately employing imperceptibly small perturbations to the benign inputs resulting in incorrect model outputs. This small perturbation is progressively amplified by a deep network and usually yields wrong predictions. Generally speaking, attacks can be a white box or black box depending on the adversary's knowledge (the agent who creates an adversarial example). White box attacks presume full knowledge of the targeted model's design, parameters, and, in some cases, training data. Gradients can thus be calculated efficiently in white box attacks using the back-propagation algorithm. In contrast, in Black box attacks, the adversary is unaware of the model parameters and has no access to the gradients. Furthermore, attacks can be targeted or un-targeted based on the intention of the adversary. Targeted attacks try to fool the model into a specific predicted output. In contrast, in the un-targeted attacks, the predicted output itself is irrelevant, and the main goal is to fool the model into any incorrect output.
% -------------------------------------------------
\begin{figure}[t]
  \captionsetup{singlelinecheck=false, font=small, belowskip=-8pt}
  \centering
    \includegraphics[width=\columnwidth]{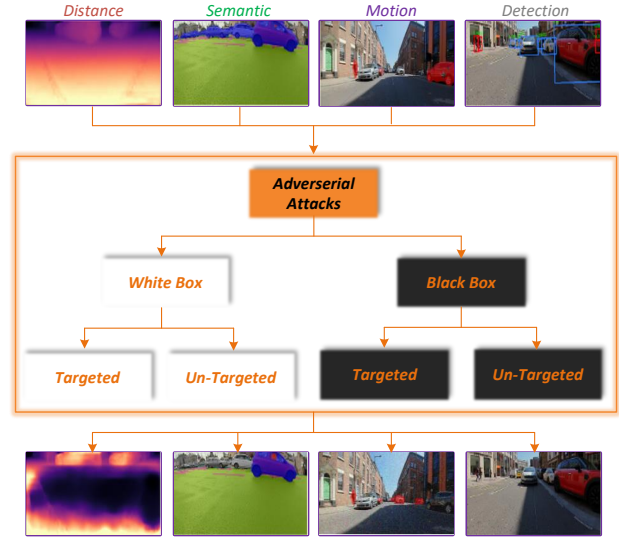}
    \caption{\textbf{Adversarial attacks on OmniDet \cite{kumar2021omnidet} MTL model.} Distance, segmentation, motion and detection perception tasks are attacked by white and black box methods with targeted and un-targeted objectives, resulting in incorrect model predictions.}
    \label{fig:teaser}
\end{figure}
% -------------------------------------------------
Fast gradient sign method (FGSM)~\cite{goodfellow2015explaining} is an example of a simple yet effective attack for generating adversarial instances. FGSM aims to fool the classification of the image by adding a small vector obtained by taking the sign of the gradient of the loss function. Moreover, it was shown that robust 3D adversarial objects could fool deep network classifiers in the physical world~\cite{athalye2018synthesizing}, despite the combination of viewpoint shifts, camera noise, and other natural transformations. Fooling surveillance cameras was introduced in~\cite{thys2019fooling} where adversarial patches are designed to attack person detection. Dense Adversary Generation (DAG) algorithm~\cite{xie2017adversarial} is an example of generating adversarial attacks for semantic segmentation and object detection tasks. It was discovered that the perturbations are exchangeable across different networks, even though they have different training since they share some intrinsic structure that makes them susceptible to a common source of perturbations. In addition to camera sensors, potential vulnerabilities of LiDAR-based autonomous driving detection systems are explored in~\cite{cao2019adversarial}. Moreover, the 3D-printed adversarial objects showed effective physical attacks on LiDAR equipped vehicles, raising concerns about autonomous vehicles' safety. Robust Physical Perturbations (RP$_2$)~\cite{eykholt2018robust} is another example that generates robust visual adversarial perturbations under different physical conditions on road sign classifications.\par

On the other hand, Adversarial robustness~\cite{kia_2021} and defense methods of neural networks have been studied to improve these networks' resistance to different adversarial attacks. One method for defense is Adversarial training, where adversarial examples besides the clean examples are both used to train the model. Adversarial training can be seen as a sort of simple data augmentation. Despite being simple, but it cannot cover all the attack cases. In~\cite{dziugaite2016study}, it is demonstrated that JPEG compression can undo the small adversarial perturbations created by the FGSM. However, this method is not effective for large perturbations. Xu~\etal~\cite{xu2018feature} proposed Feature-squeezing for detecting adversarial examples, in which the model is tested on both original input and the input after being pre-processed by feature squeezers such as spatial smoothing. If the difference between the outputs exceeds a certain threshold, we identify the input as an adversarial example. Defense-GAN~\cite{samangouei2018defense} is another defence technique that employs generative adversarial networks (GAN)s~\cite{goodfellow2014generative}, in which it seeks a similar output to a given picture while ignoring adversarial perturbations. It is shown to be a feasible defense that relies on the GAN's expressiveness and generative power. However, training GANs is still a challenging task. 

Robust attacks and defenses are still challenging tasks and an active area of research. Most previous works on adversarial attacks focused on single task scenarios. However, in real-life situations, multi-task learning is adopted to solve several tasks at once~\cite{sistu2019real, Chennupati_2019, kumar2021syndistnet}. Accordingly, multi-task networks are used to leverage the shared knowledge among tasks, leading to better performance, reduced storage, and faster inference~\cite{leang2020dynamic, kumar2021svdistnet}. Moreover, it is shown that when models are trained on multiple tasks at once, they become more robust to adversarial attacks on individual tasks~\cite{mao2020multitask}. However, defense remains an open challenge. 
In this work, as shown in Figure~\ref{fig:teaser}, white and black box attacks are applied on a multi-task visual perception deep network across distance estimation, semantic segmentation, motion detection, and object detection, taking into consideration both targeted and un-targeted scenarios. For the experiment, while attacking one of the tasks, the attacking curve is plotted to inspect the performance across all the tasks over the attacking steps. Additionally, a simple defense method is used across all experiments. Finally, in addition to visual samples of perturbations and performance before and after the attacks, detailed results and comparisons are presented and discussed.\par
% -------------------------------------------------
% -------------------------------------------------
% -------------------------------------------------'
\section{Multi-Task Adversarial Attacks}

In this section, the target multi-task network is presented in terms of architecture, data, tasks, and training. Then the attacks are detailed for each task, including a white and black box for both targeted and un-targeted cases.\par
% -------------------------------------------------
\subsection{Baseline Multitask model}

\begin{figure}[t]
  \captionsetup{singlelinecheck=false, font=small, belowskip=-8pt}
  \centering
    \includegraphics[width=\columnwidth]{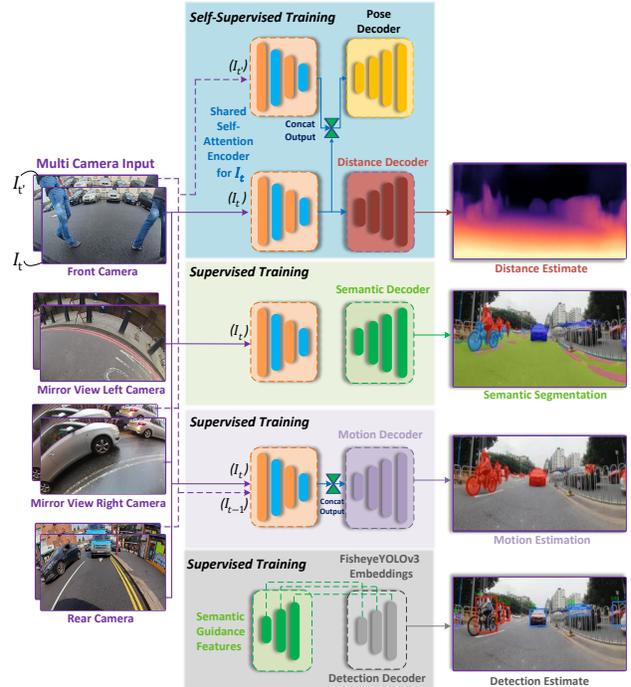}
    \caption{\textbf{Illustration of baseline multi-task architecture comprising of four tasks \cite{kumar2021omnidet}.}}
    \label{fig:mtl_pipeline}
\end{figure}

We derive the baseline model from the recent work OmniDet~\cite{kumar2021omnidet, kumar2020fisheyedistancenet, kumar2020unrectdepthnet, kumar2021fisheyedistancenet++}, a six-task complete perception model for surround view fisheye cameras. We focus on the four main perception tasks and skip visual odometry and the soiling detection task. We provide a short overview of the baseline model used and refer to \cite{kumar2021omnidet} for more details.
A high-level architecture of the model is shown in Figure \ref{fig:mtl_pipeline}. It comprises a shared ResNet18 encoder and four decoders for each task. Motion decoder uses additionally previous encoder feature in siamese encoder style.
2D box detection task has the five important objects, namely \textit{pedestrians, vehicles, riders, traffic sign, and traffic lights}.
Segmentation task has \textit{vehicles, pedestrians, cyclists, road, lanes, and curbs} categories. Motion task has binary segmentation corresponding to static and moving \ie dynamic pixels. Depth task provides scale-aware distance in 3D space validated by occlusion corrected LiDAR depth \cite{kumar2018near, kumar2018monocular}. The model is trained jointly using the public WoodScape~\cite{yogamani2019woodscape} dataset comprising 8k samples and evaluated on 2k samples.\par
In this paragraph, we briefly summarize the loss functions used for training. We construct a self-supervised monocular structure-from-motion (SfM) system for distance and pose estimation. The total loss consists of a photometric term $\mathcal{L}_r$, a smoothness term $\mathcal{L}_s$, that enforces edge-aware smoothness within the distance map $\hat{D}_t$, a cross-sequence distance consistency loss $\mathcal{L}_{dc}$, and feature-metric losses from~\cite{shu2020feature} where $\mathcal{L}_{dis}$ and $\mathcal{L}_{cvt}$ are computed on $I_t$. Final loss function for distance estimation is weighted average of all these losses. The segmentation task contains seven classes on the WoodScape and employs \textit{Lovasz-Softmax}~\cite{berman2018lovasz} loss. Motion segmentation employs two frames and predicts either a binary moving or static mask and employs \textit{Lovasz-Softmax}~\cite{berman2018lovasz}, and \textit{Focal}~\cite{lin2017focal} loss for managing class imbalance instead of the cross-entropy loss. For object detection, we make use of YOLOv3 loss and add IoU loss using segmentation mask \cite{rashedfisheyeyolo, rashed2021generalized}.\par
% -------------------------------------------------
% -------------------------------------------------
\subsection{Experimental setup for Attacks}

In this section, the details of the experiments are described. We conduct the experiments across the four visual perception tasks, on a test set of 100 images \ie randomly sampled from the original test set of the target network. We generate the Adversarial examples for each image in the test set while attacking one task at a time. 

We make use of the standard FGSM method \cite{goodfellow2015explaining} for the white box attack. 
Using the gradients of the network, we perform an \textit{iterative optimization process} to add perturbation in the input image in a direction to harm the original predictions. For Black box attacks we set up similar protocols established in the white-box; however, the gradients are not given but estimated. As a generic black-box optimization algorithm, we show that Evolution Strategies (ES) can be adopted as a black-box optimization method for generating adversarial examples. Precisely, (ES) algorithm is used to update the adversarial example over the attacking steps. At each step, we take the adversarial example vector, \ie the image, and generate a population of 25 slightly different vectors by adding noise sampled from a normal distribution. Then, we evaluate each of the individuals by feeding it through the network. Finally, the newly updated vector is the weighted sum of the population vectors. Each weight is proportional to the task's desired performance, and the process continues till convergence or stops criteria.\par

In the un-targeted case, the aim is to harm the predictions the most without considering a certain target prediction: $f(x_{adv}) \neq y_{true}$, however in the targeted case, the aim is to harm the predictions in a desired specific way towards a certain target: $f(x_{adv}) = y_{target}$. The attack loss is based on the task. Mean square error (MSE) is used for the distance task, while cross-entropy loss is used for motion and semantic segmentation tasks. For the object detection task, only object confidence is attacked, and hence cross-entropy loss is adopted. Regarding un-targeted attacks across all the tasks, the goal is to maximize the distance between the original output of the network and the adversarial example's output. Accordingly, we add the perturbations to achieve this simple goal where the output can be anything but the correct one. This can be formulated as $\theta = \theta + \alpha dJ/d\theta$ where $\theta$ is the image parameters \ie pixels, $J$ the loss functions and $\alpha$ is the learning rate. However, for the targeted attacks, the target output is defined. The aim is to minimize the distance between the original output and the target output according to $\theta = \theta - \alpha dJ/d\theta$.\par

For each perception task, the targets are as follows: \textit{Targeted Depth} attack tries to convert the predicted near pixels to be predicted as far. The \textit{Targeted Segmentation} attack tries to convert the predicted vehicle pixels as void for randomly 50\% of the test set. For the other 50\%, tries to convert the predicted road pixels as void. The \textit{Targeted Motion} attack tries to convert the predicted dynamic object pixels to be predicted as static. Finally, similar to semantic segmentation, the \textit{Targeted Object Detection} attack tries to increase or decrease the predicted confidence randomly. In addition to attacks, we apply a simple blurring defense approach across all the attacks. Similar to~\cite{dziugaite2016study}, the intuition is trying to remove the adversarial perturbations and restore the original output as much as possible. 
The hyperparameters of the attacks are empirically defined based on a very small validation set of three samples. All white-box attacks are conducted with learning rate $\alpha = 0.00015$. In black-box attacks, the hyperparameters are chosen to balance the attack effect and the severity of the perturbations, where the learning rates range from $0.0001$ to $0.001$, and $\mu=0, \sigma = 0.05$ for ES population generation.\par
% -------------------------------------------------
% -------------------------------------------------
% -------------------------------------------------
\section{Results} 

In this section, we present and discuss the details of the results. As expected, white-box attacks, with the gradients are accessible, were easier to find than the black box case. White box attacks can generate adversarial examples with minimal and localized perturbations across all the tasks. On the other hand, ES black-box attacks have more significant perturbations and require more hyperparameters to optimize.\par

\begin{figure*}[t]
    \centering
    
    \includegraphics[width=0.45\textwidth, trim={2.35cm 1.4cm 2.55cm 1.6cm},clip]{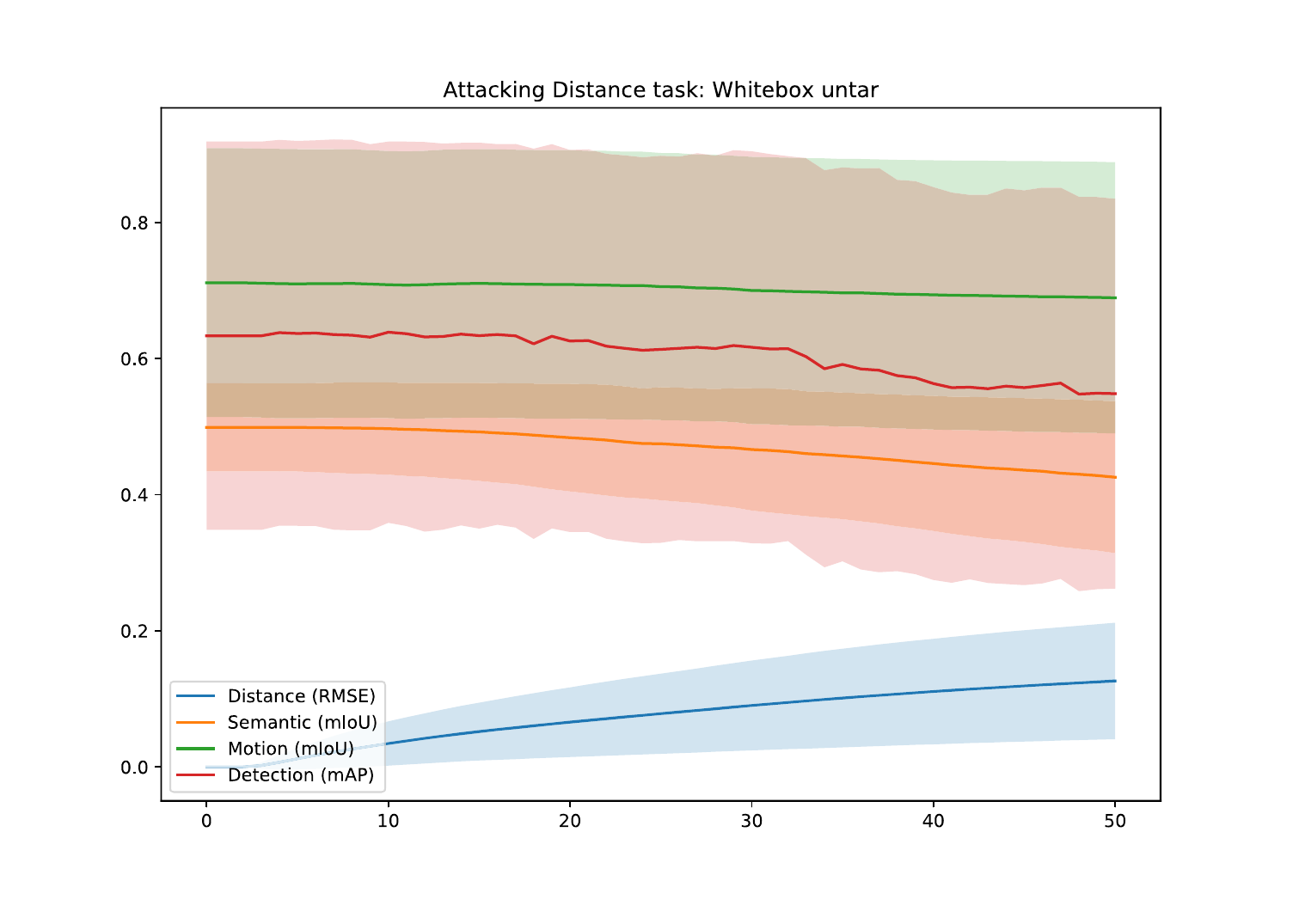}
    \includegraphics[width=0.45\textwidth, trim={2.35cm 1.4cm 2.55cm 1.6cm},clip]{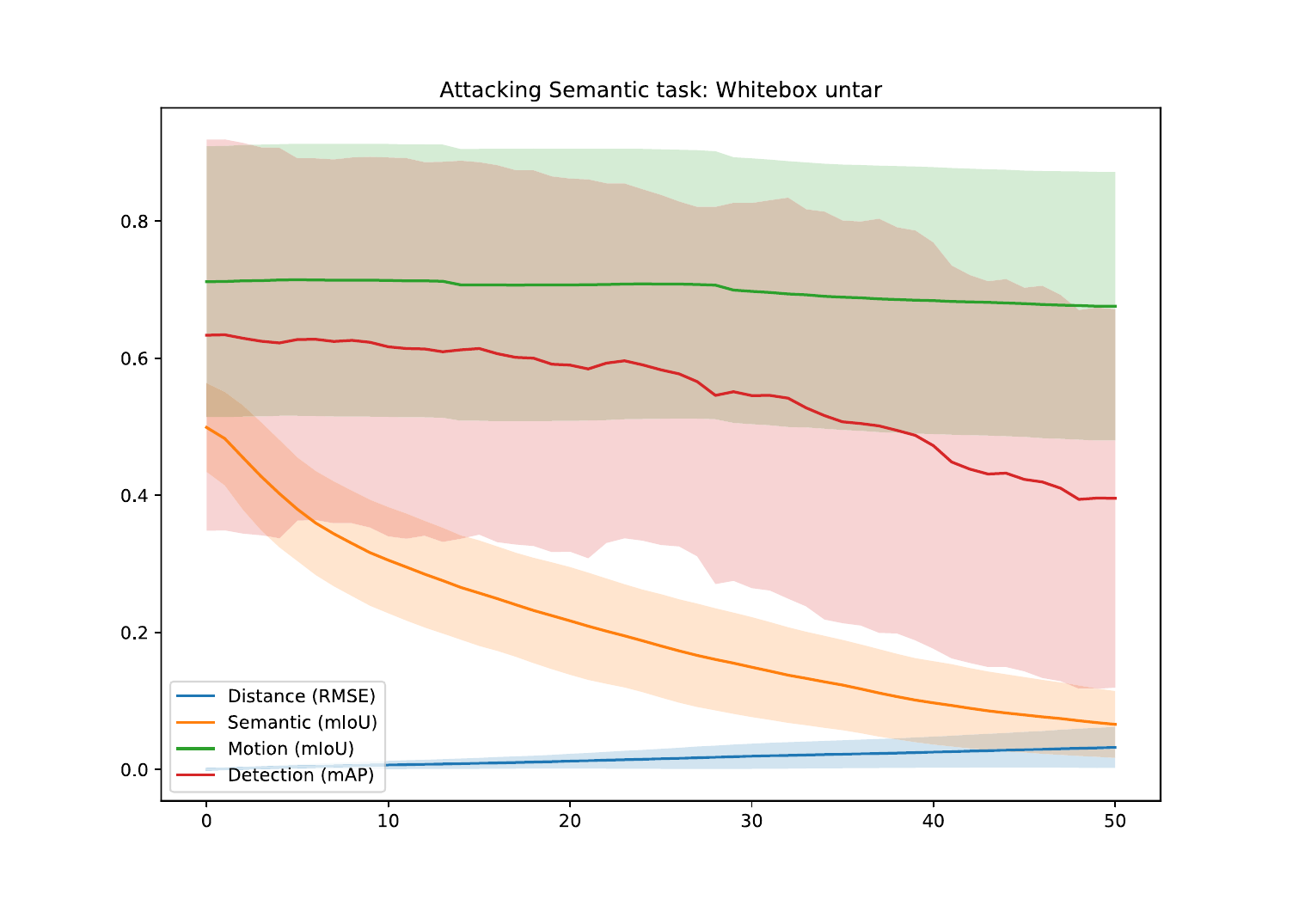}  
    \includegraphics[width=0.45\textwidth, trim={2.35cm 1.4cm 2.55cm 1.6cm},clip]{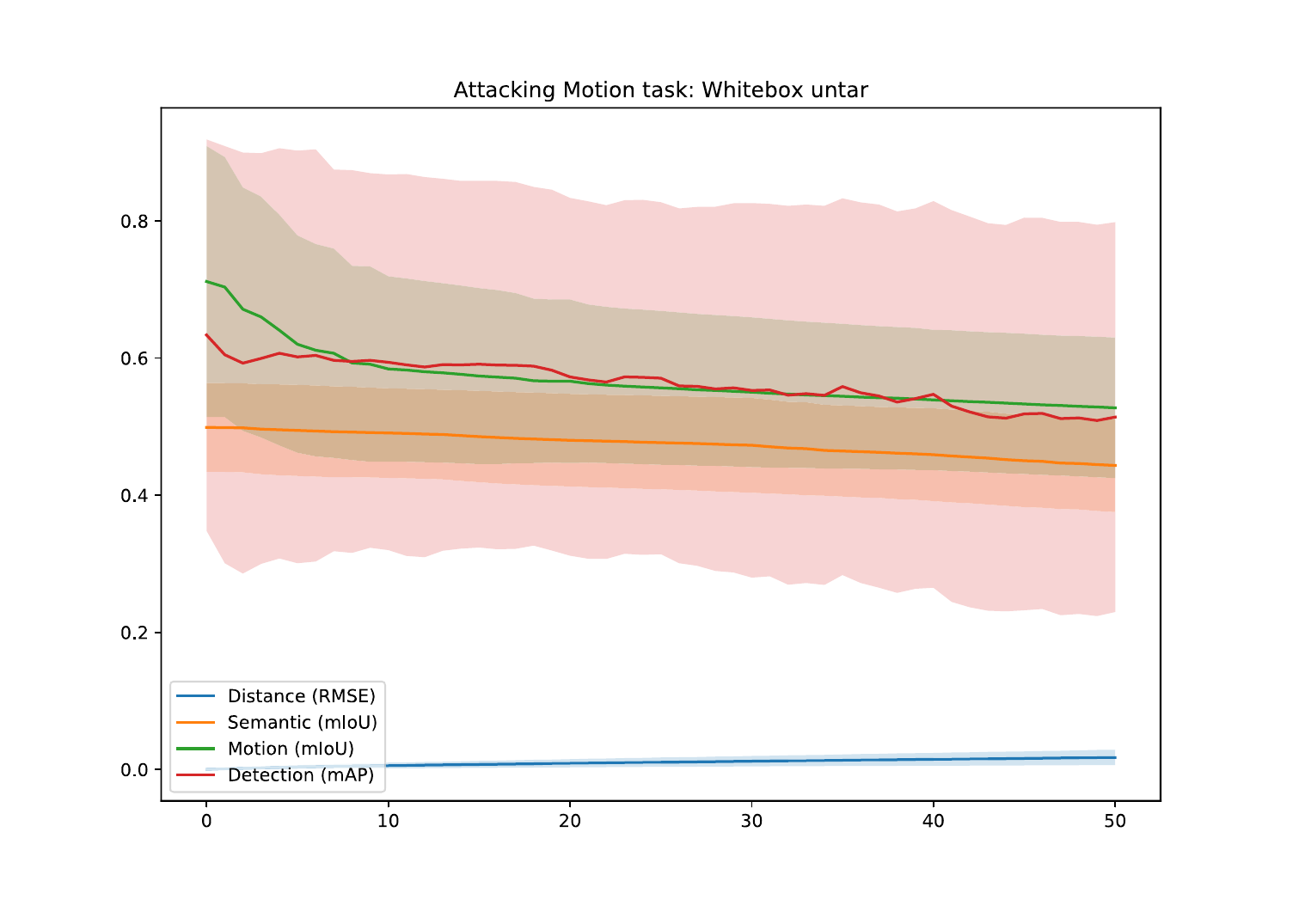}
    \includegraphics[width=0.45\textwidth, trim={2.35cm 1.4cm 2.55cm 1.6cm},clip]{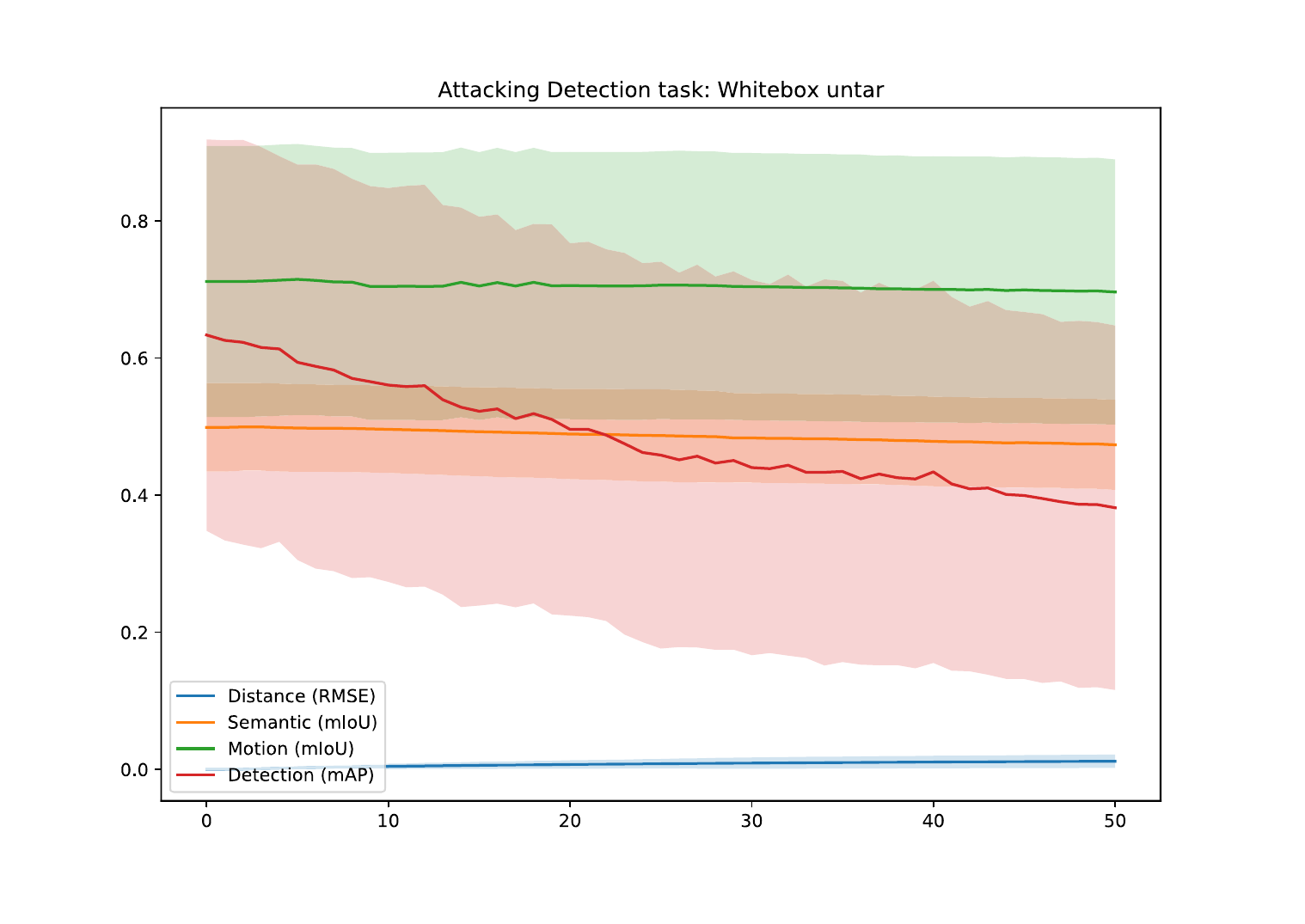}   
\\
\vspace{1mm}
    \includegraphics[width=0.45\textwidth, trim={2.35cm 1.4cm 2.55cm 1.6cm},clip]{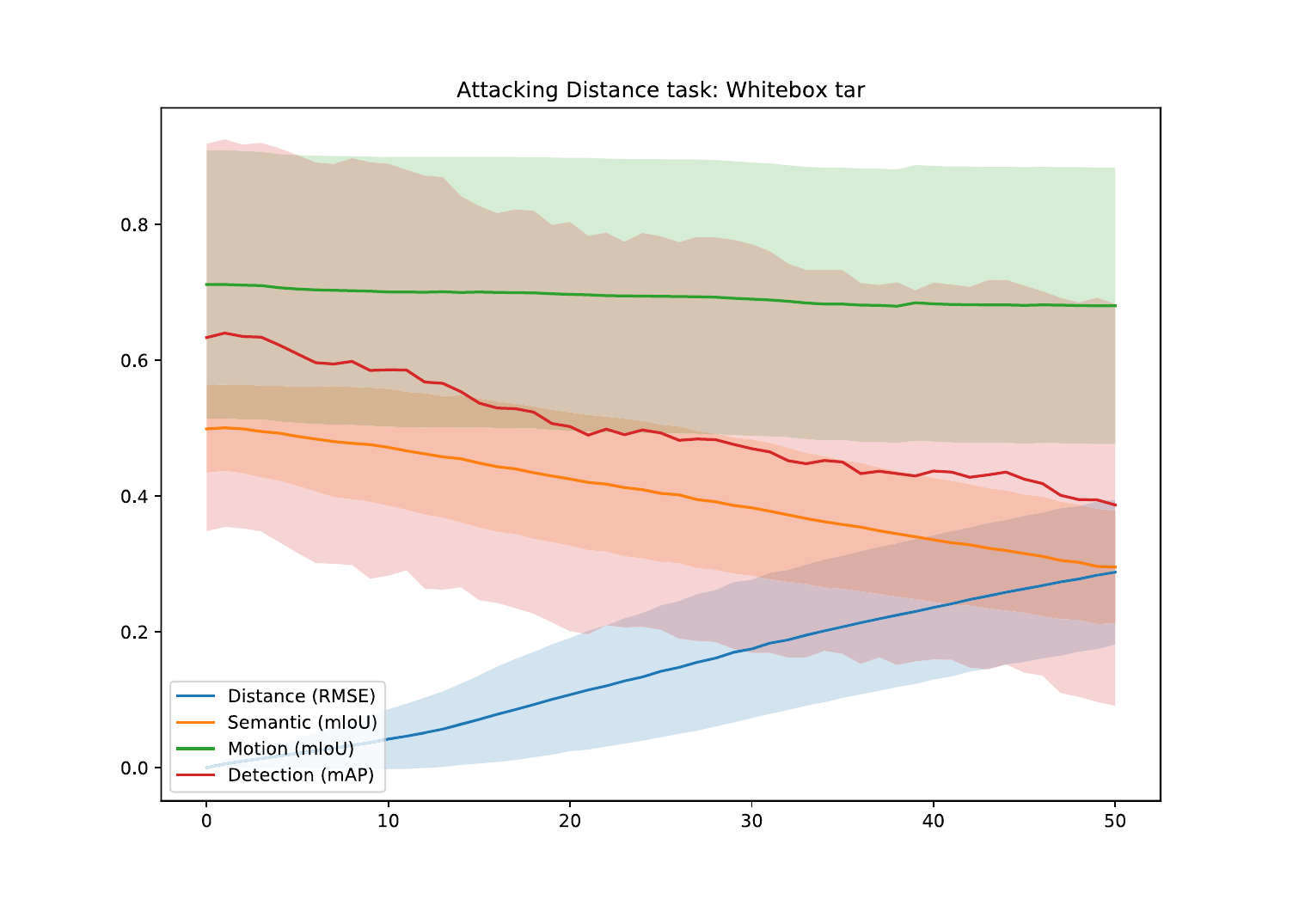}
    \includegraphics[width=0.45\textwidth, trim={2.35cm 1.4cm 2.55cm 1.6cm},clip]{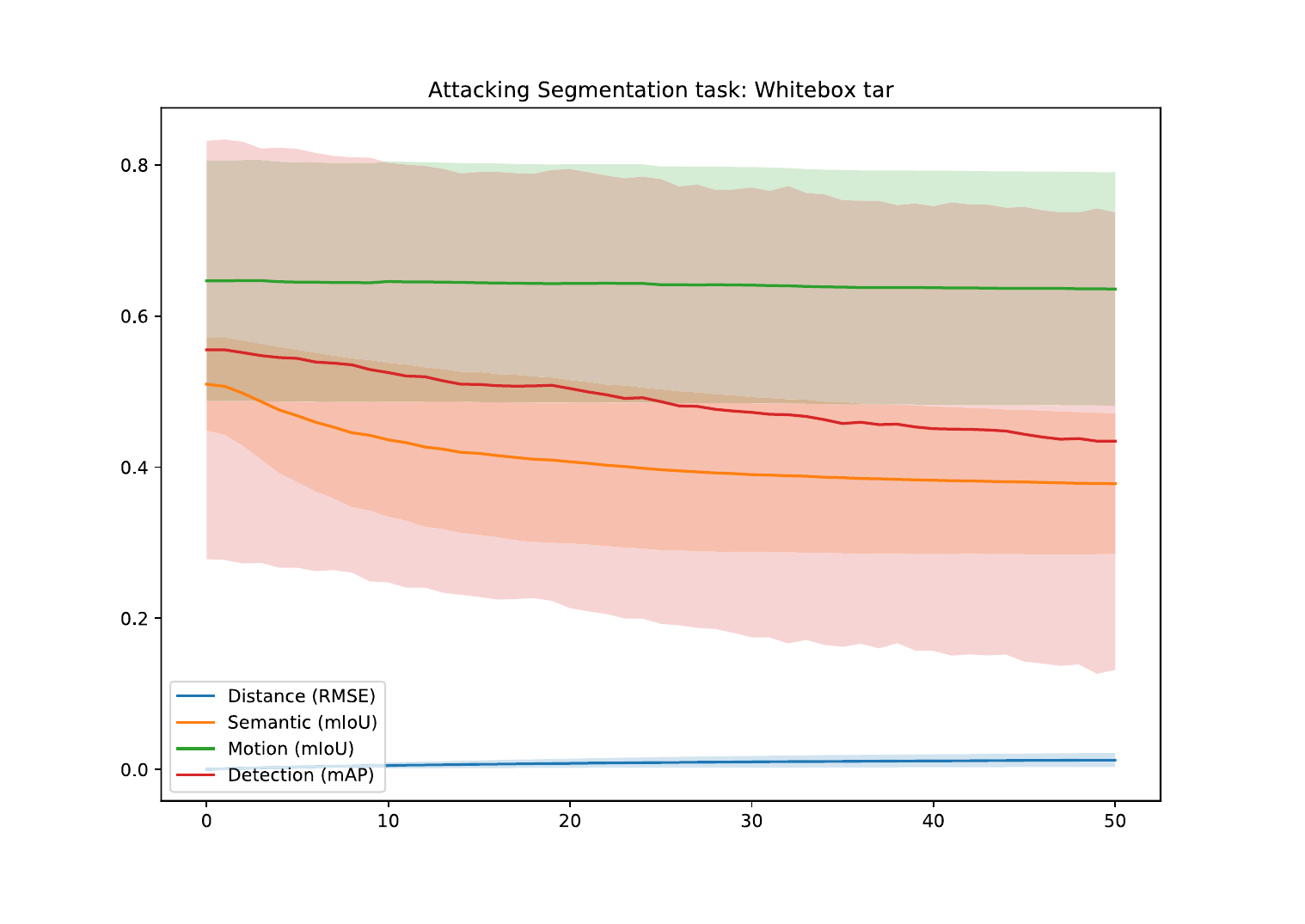}  
    \includegraphics[width=0.45\textwidth, trim={2.35cm 1.4cm 2.55cm 1.6cm},clip]{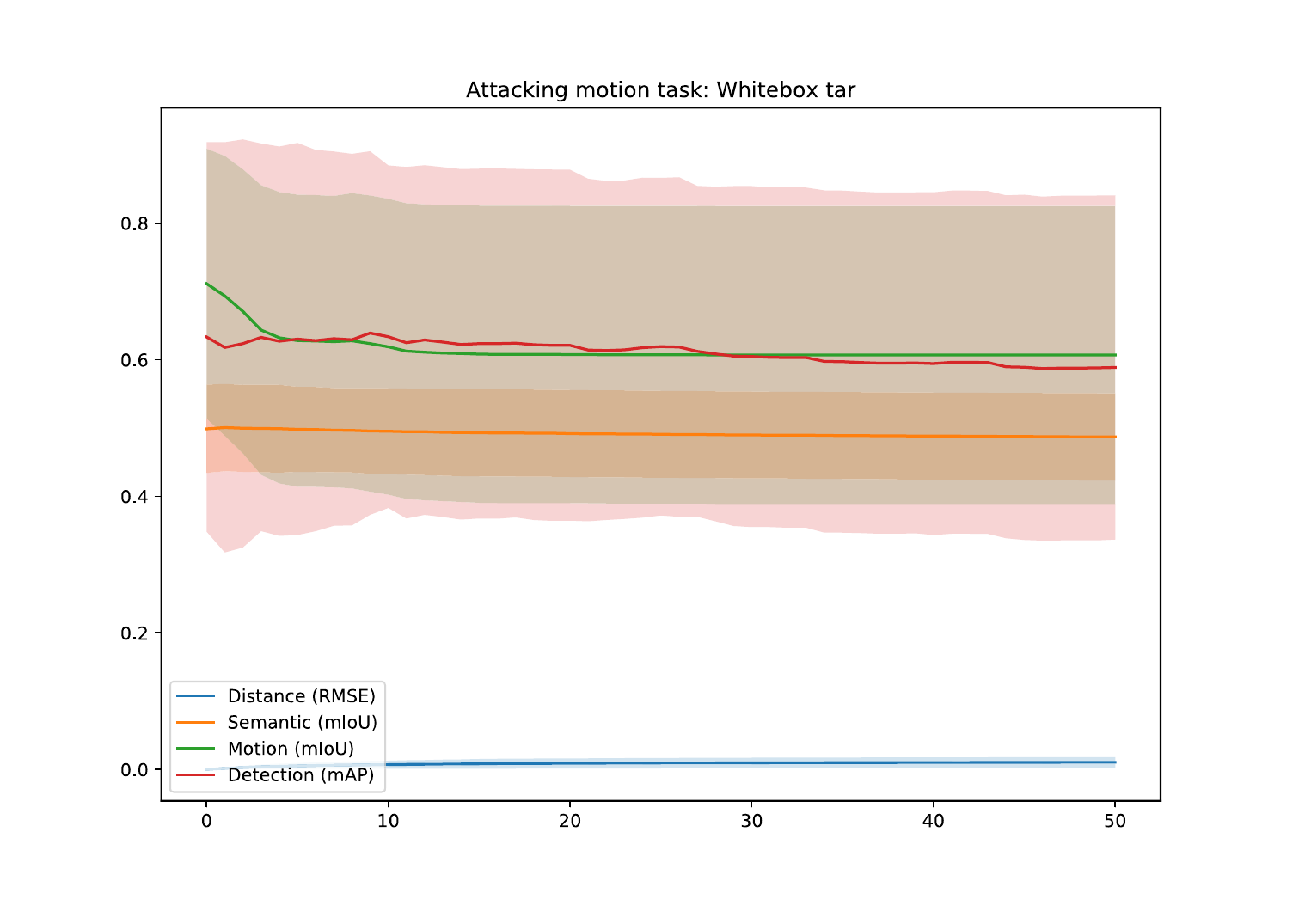}
    \includegraphics[width=0.45\textwidth, trim={2.35cm 1.4cm 2.55cm 1.6cm},clip]{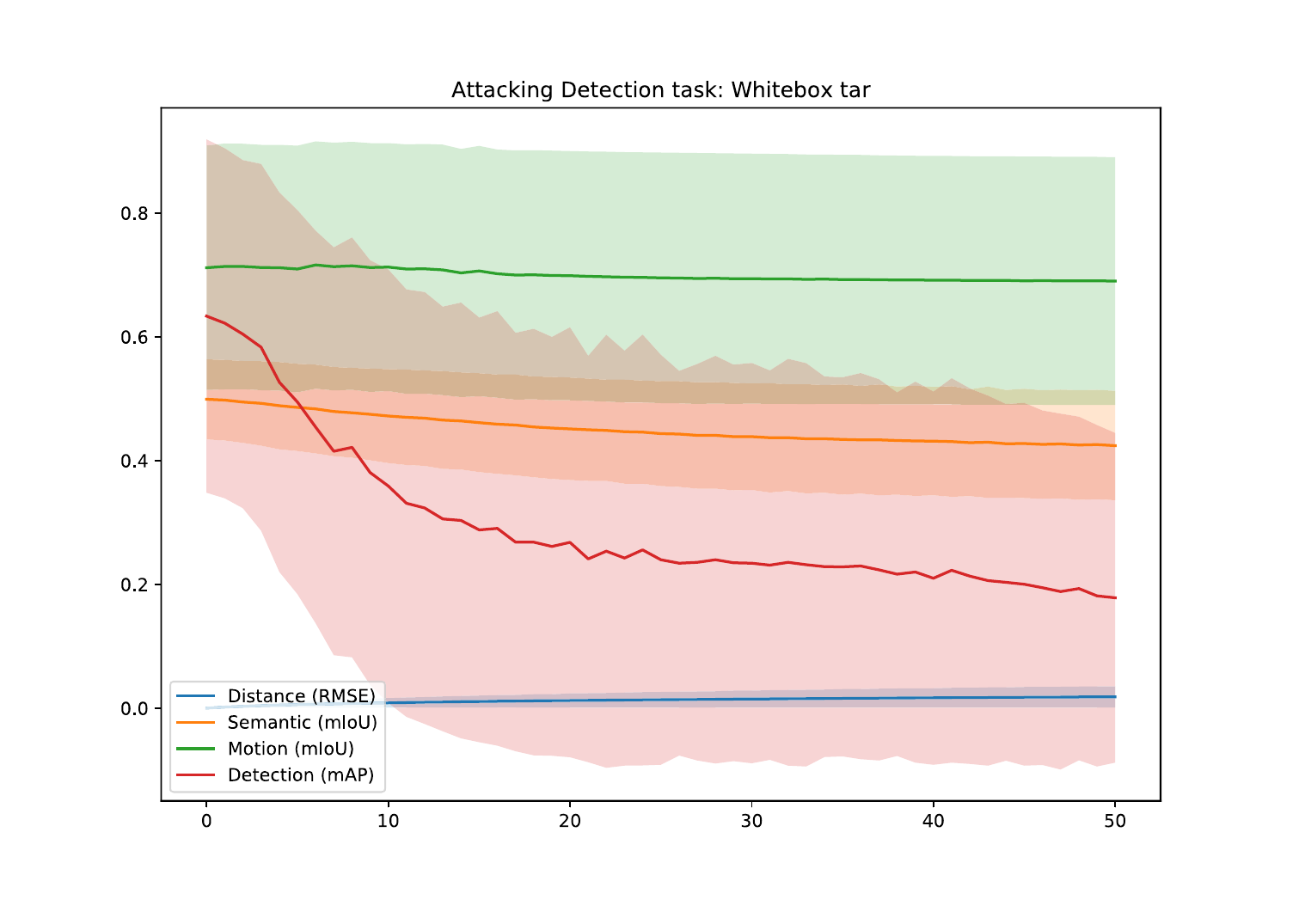}  

    \caption{Performance comparison of the \textbf{White box} attacks across different tasks. The first row shows the un-targeted attacks, the second row shows the targeted attacks, and columns represent the tasks.}
    \label{fig:whitebox}
\end{figure*}

\begin{figure*}[t]
    \centering
    \includegraphics[width=0.45\textwidth, trim={2.35cm 1.4cm 2.55cm 1.6cm},clip]{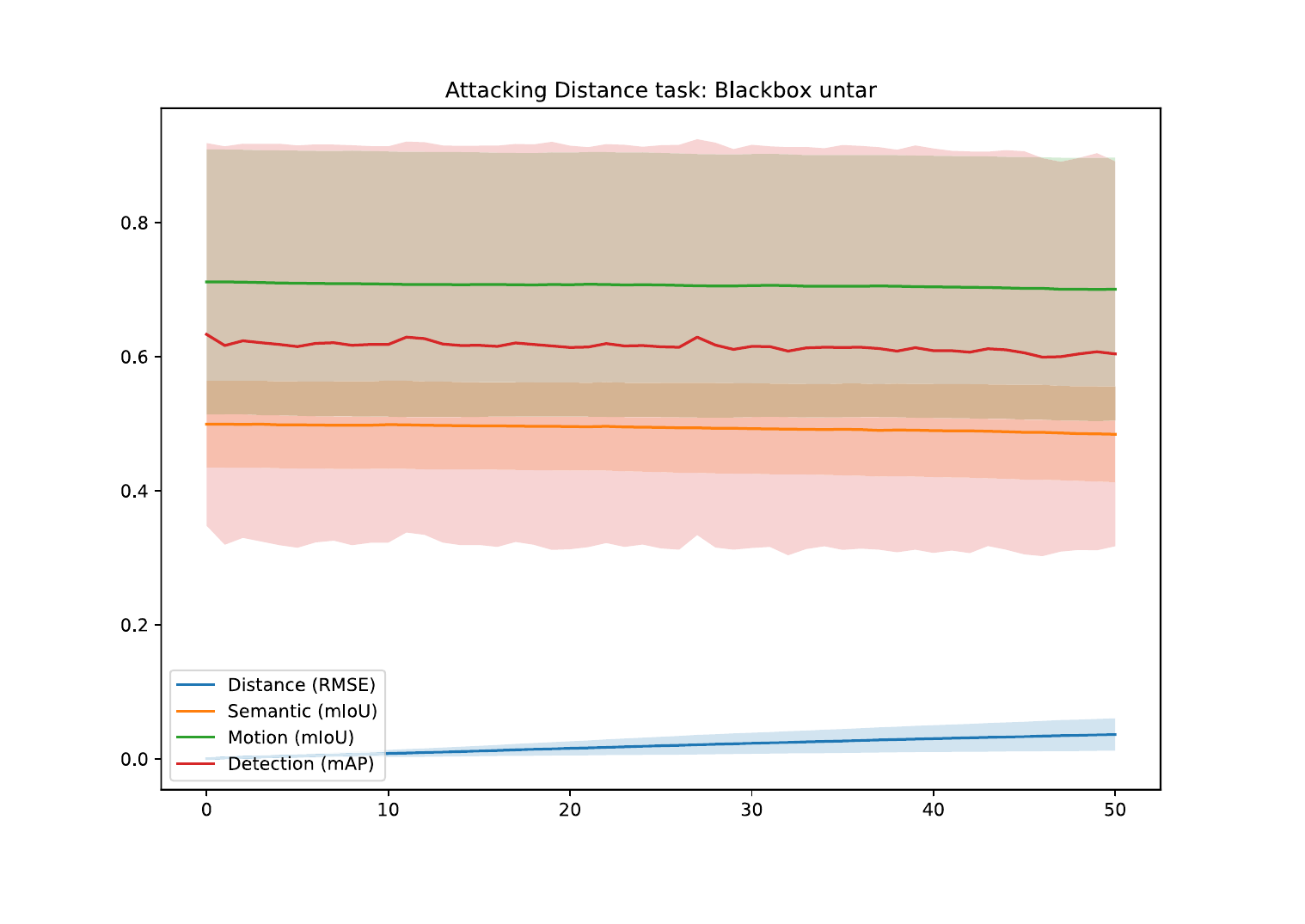}
    \includegraphics[width=0.45\textwidth, trim={2.35cm 1.4cm 2.55cm 1.6cm},clip]{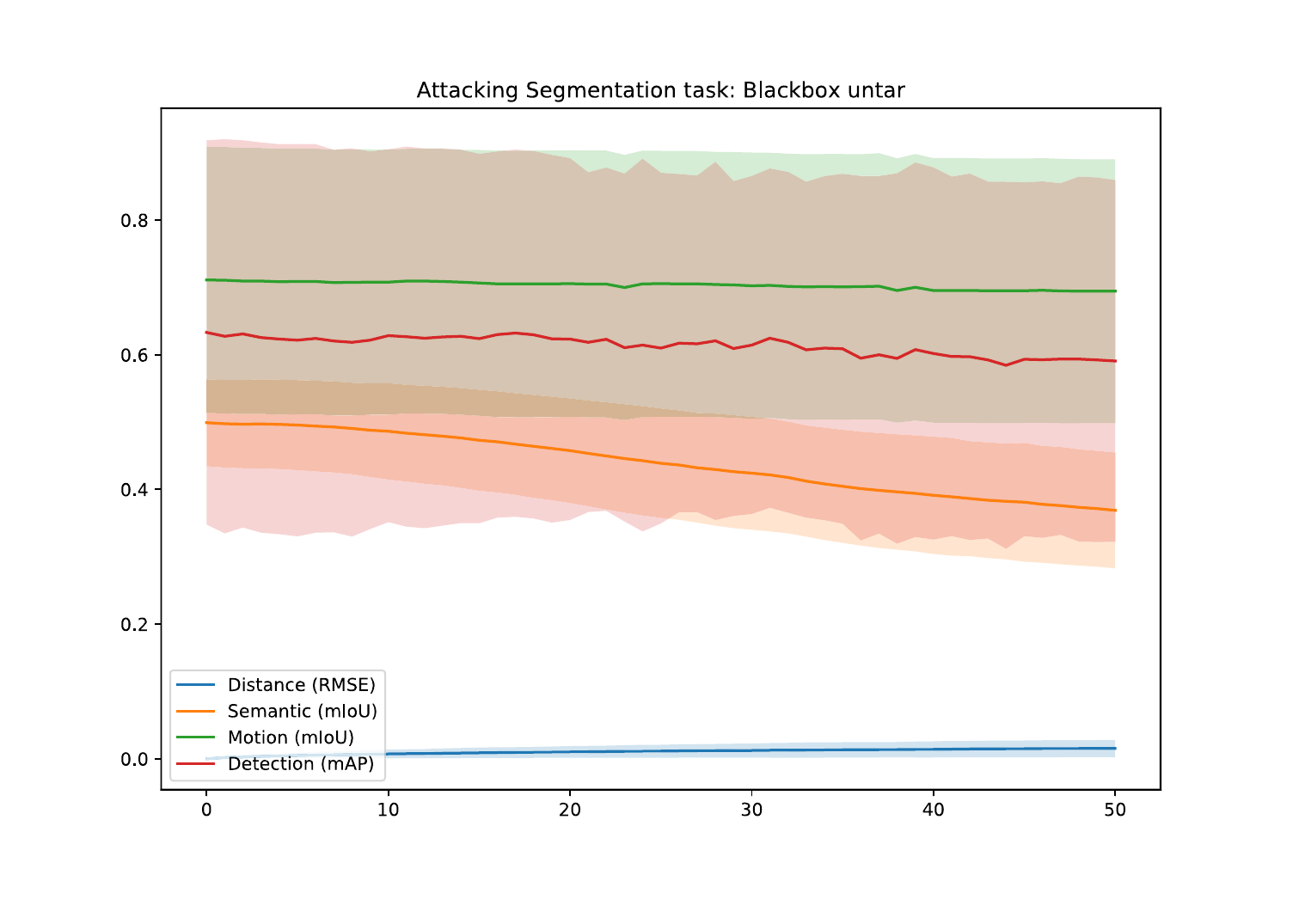}  
    \includegraphics[width=0.45\textwidth, trim={2.35cm 1.4cm 2.55cm 1.6cm},clip]{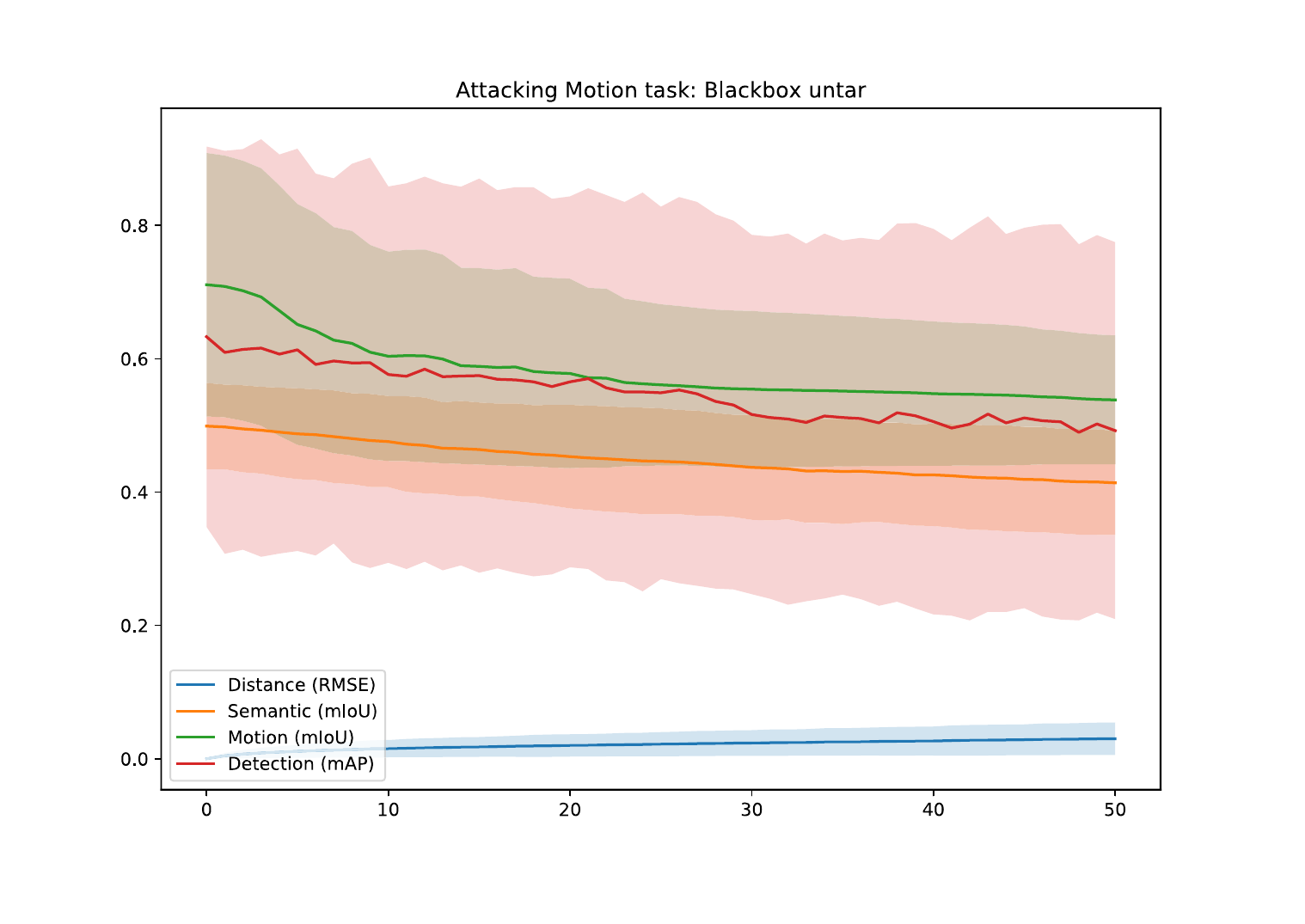}
    \includegraphics[width=0.45\textwidth, trim={2.35cm 1.4cm 2.55cm 1.6cm},clip]{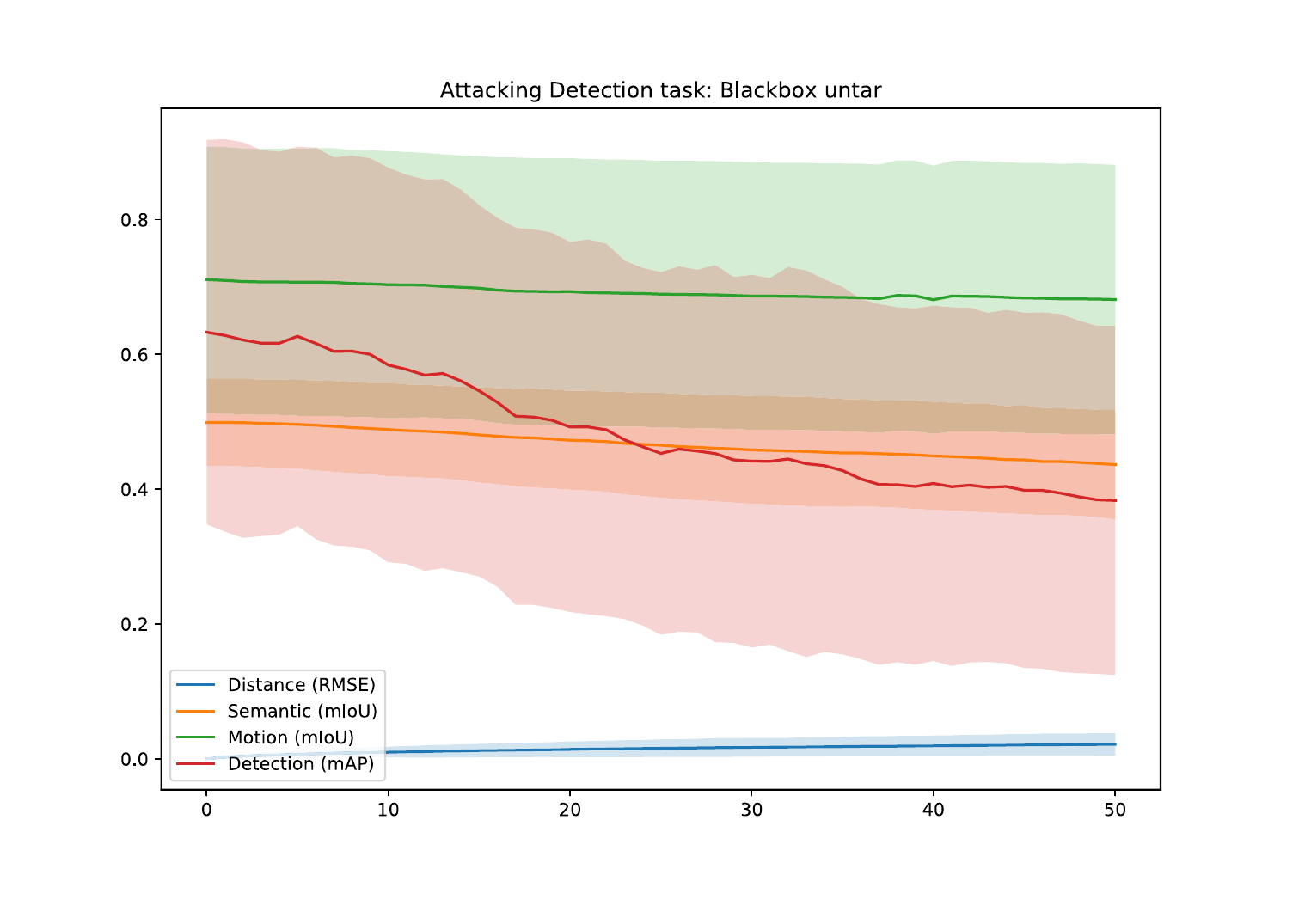}  
\\
\vspace{1mm}
    \includegraphics[width=0.45\textwidth, trim={2.35cm 1.4cm 2.55cm 1.6cm},clip]{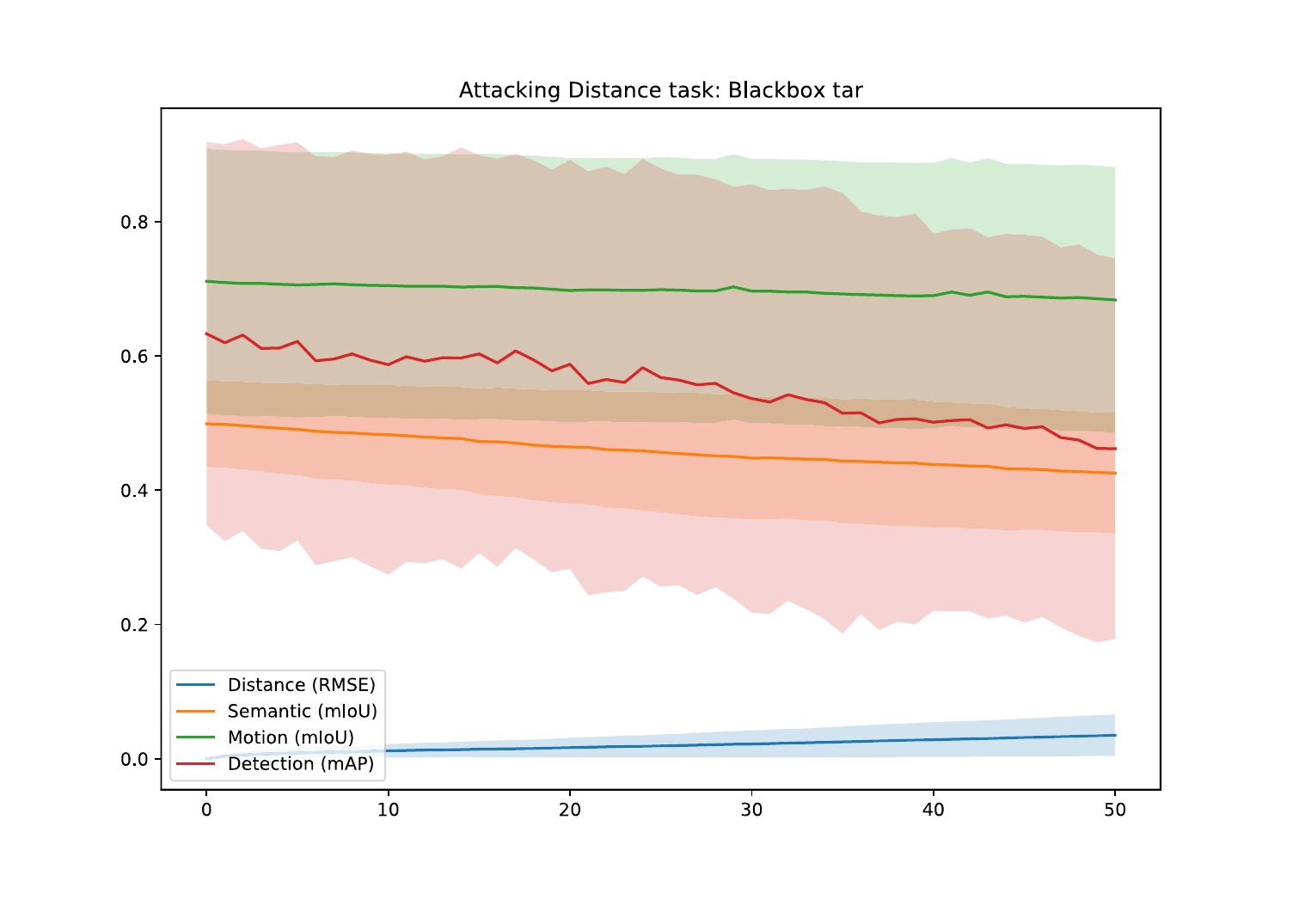}
    \includegraphics[width=0.45\textwidth, trim={2.35cm 1.4cm 2.55cm 1.6cm},clip]{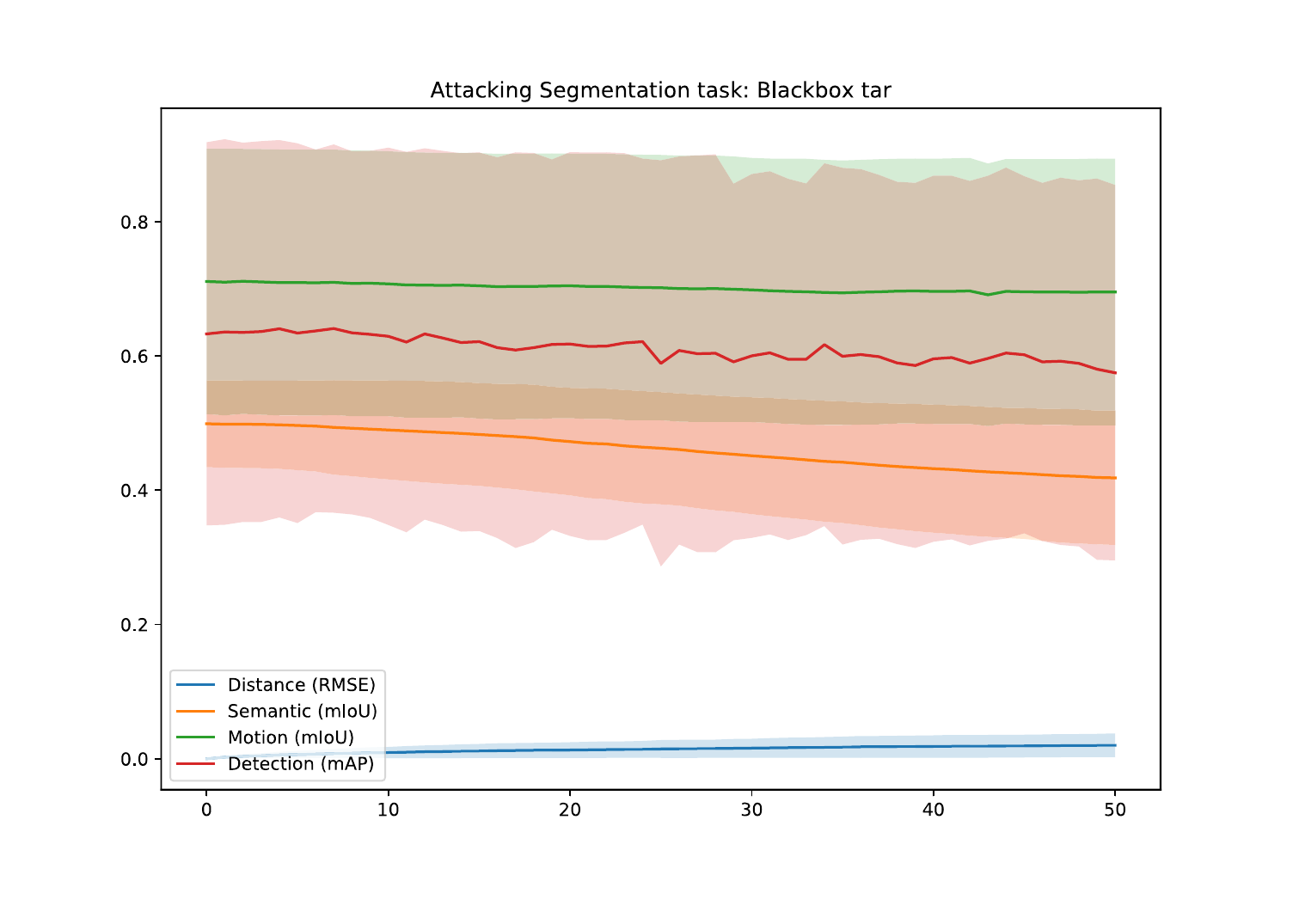}  
    \includegraphics[width=0.45\textwidth, trim={2.35cm 1.4cm 2.55cm 1.6cm},clip]{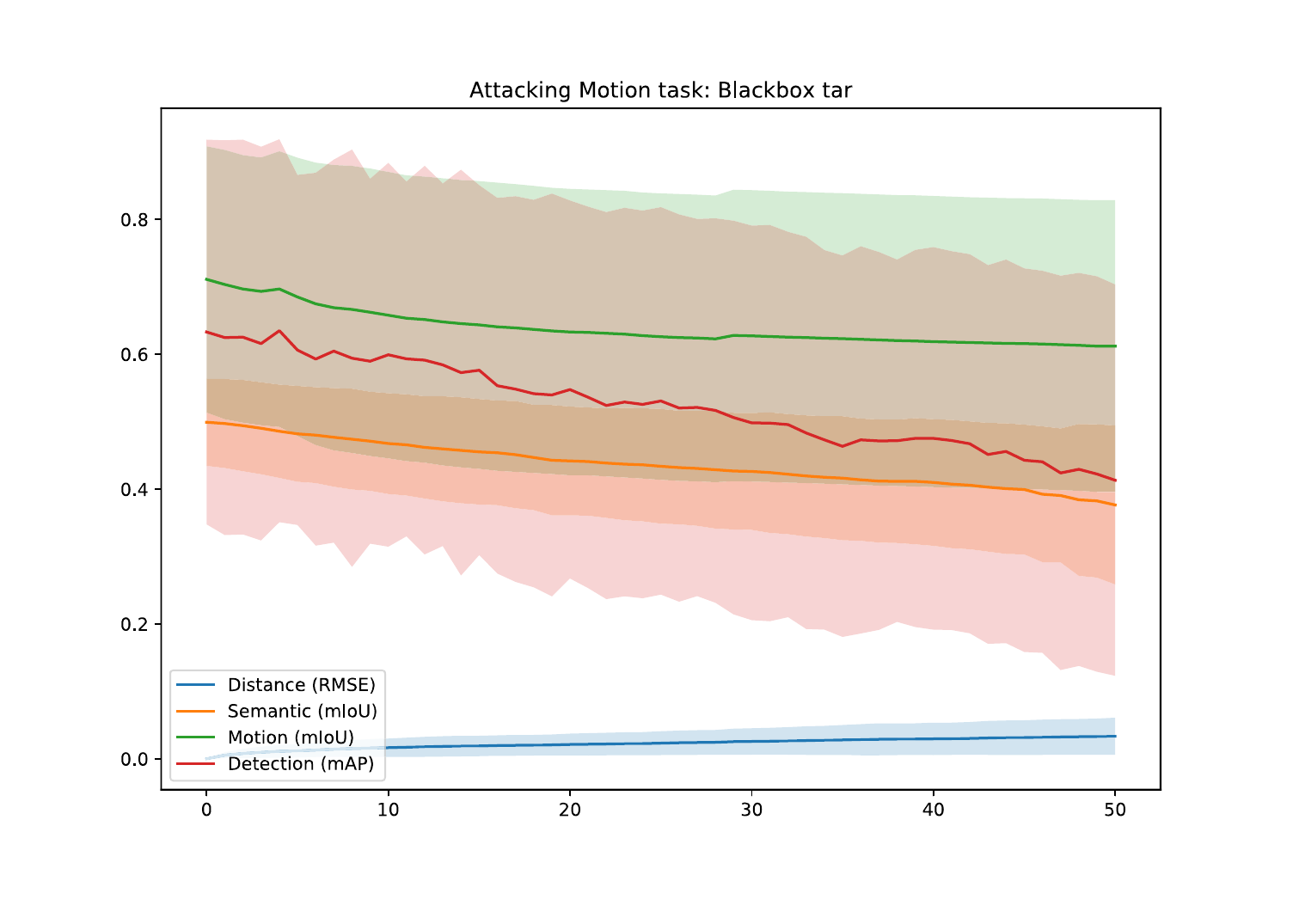}
    \includegraphics[width=0.45\textwidth, trim={2.35cm 1.4cm 2.55cm 1.6cm},clip]{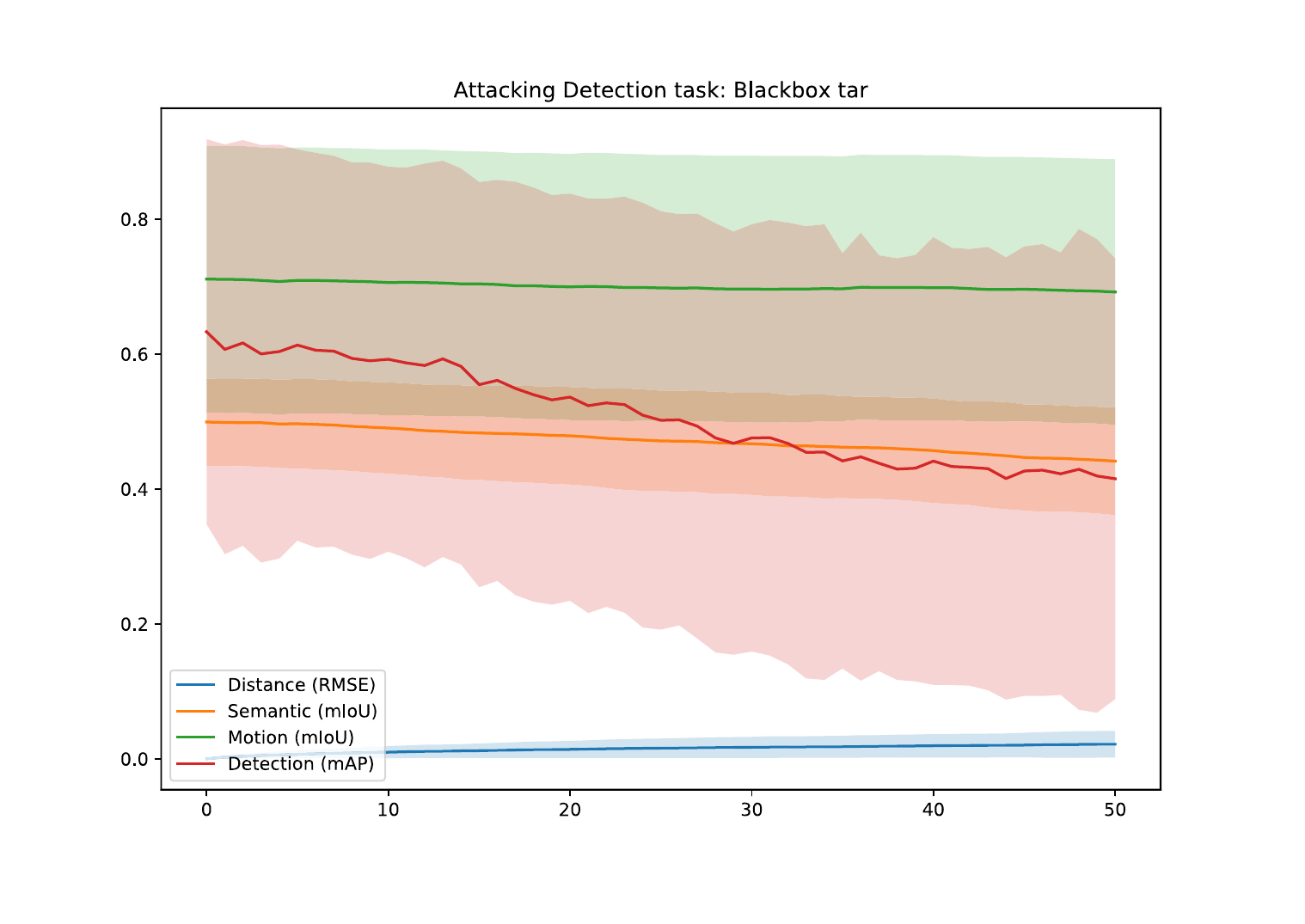}  
    
\caption{Performance comparison of the \textbf{Black box} attacks across different tasks. The first row shows the un-targeted attacks, the second row shows the targeted attacks, and columns represent the tasks.}
\label{fig:blackbox}    
\end{figure*}

The attacking curves for white and black box attacks are shown in Figures \ref{fig:whitebox} and \ref{fig:blackbox} respectively. Each plot shows each perception task's performance over the 50 attacking steps where the first step at index $0$ represents the actual performance of the target network without applying any attack. Each curve shows the mean performance of a task over the test set, where the shaded area is the mean $\pm$ standard deviation. Generally, Motion and detection tasks have a performance with a large standard deviation indicating the test set's diversity containing easy and hard examples. Across all curves, it is clear that the performance is decreasing along with the attacking steps.\par

Moreover, attacking one task by generating an adversarial example affects the other tasks' performance in different ways along the attacking curve. These curves enable the adversary to decide at which step the adversarial example is generated according to the required effect on the target task and the other tasks. As shown in Figures \ref{fig:whitebox} and \ref{fig:blackbox}, in most cases, attacking other tasks has a marginal negative effect on motion task. The main reason is that the motion task takes two frames as input, and only one of them is attacked. Moreover, it is shown that the attacking distance task affects both segmentation and detection tasks. Attacking segmentation or detection showed to affect other tasks. As mentioned, the attack effect depends on the parameters selected for the attack. Moreover, targeted attacks try to optimize the adversarial example to produce the required target prediction. In contrast, the un-targeted attack continues to apply perturbation to produce as different as possible predictions.\par
\begin{figure*}[t]
    \centering
%     \includegraphics[width=0.245\textwidth,clip]{samples/W213-263.YUV.4CAM.DAT.20180830.141850_FV_601/cropped.png}
%     \includegraphics[width=0.245\textwidth ,clip]{samples/W213-263.YUV.4CAM.DAT.20180525.131730_MVR_241/cropped.png}  
%     \includegraphics[width=0.245\textwidth ,clip]{samples/W213-263.YUV.4CAM.DAT.20180820.123958_RV_571/cropped.png}
%     \includegraphics[width=0.245\textwidth ,clip]{samples/W213-263.YUV.4CAM.DAT.20180822.101741_MVL_421/cropped.png}   
    
% \vspace{.5mm}
%   \includegraphics[width=0.245\textwidth ,clip]{samples/W213-263.YUV.4CAM.DAT.20180830.141850_FV_601/adv_x_wb_depth_untar.png}
%     \includegraphics[width=0.245\textwidth ,clip]{samples/W213-263.YUV.4CAM.DAT.20180525.131730_MVR_241/adv_x_wb_ss_untar.png}  
%     \includegraphics[width=0.245\textwidth ,clip]{samples/W213-263.YUV.4CAM.DAT.20180820.123958_RV_571/adv_x_wb_motion_untar.png}
%     \includegraphics[width=0.245\textwidth ,clip]{samples/W213-263.YUV.4CAM.DAT.20180822.101741_MVL_421/adv_x_wb_od_untar.png} 

%\vspace{.5mm}
    \includegraphics[width=0.245\textwidth ,clip]{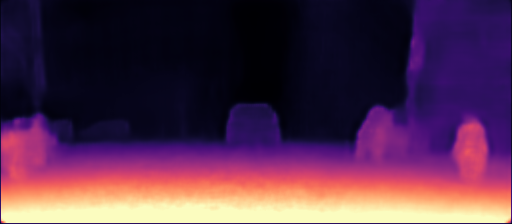}
    \includegraphics[width=0.245\textwidth, clip]{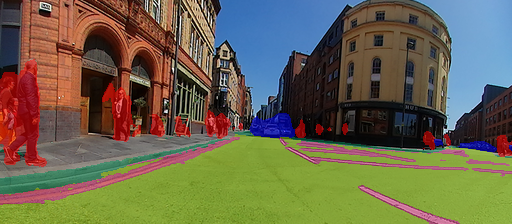}  
    \includegraphics[width=0.245\textwidth, clip]{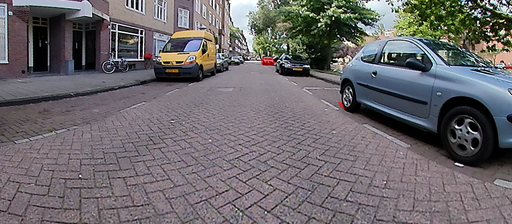}
    \includegraphics[width=0.245\textwidth, clip]{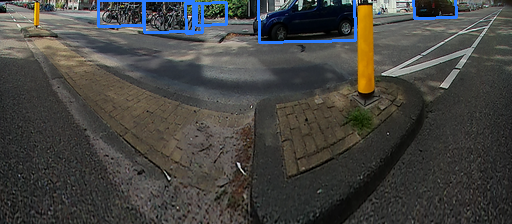}  
    
% \vspace{.5mm}
    \includegraphics[width=0.245\textwidth ,clip]{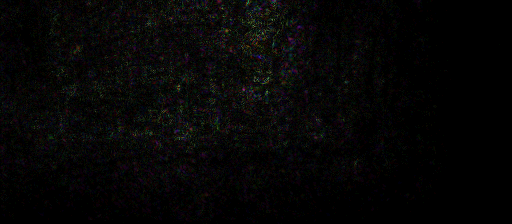}
    \includegraphics[width=0.245\textwidth ,clip]{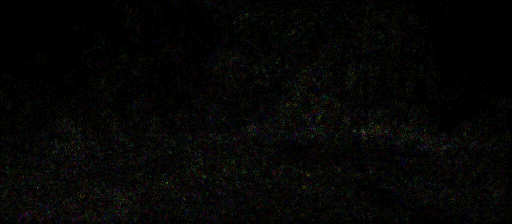}  
    \includegraphics[width=0.245\textwidth ,clip]{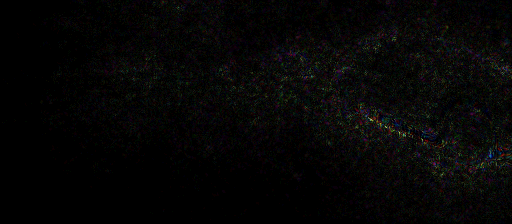}
    \includegraphics[width=0.245\textwidth, clip]{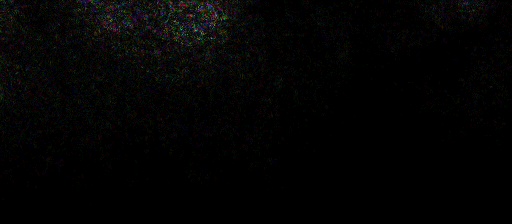} 
    
%\vspace{.5mm}
    \includegraphics[width=0.245\textwidth, clip]{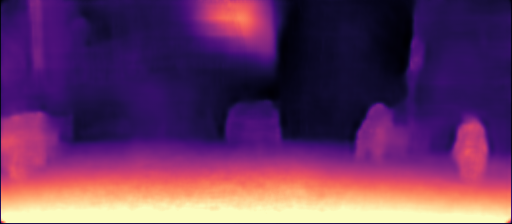}
    \includegraphics[width=0.245\textwidth,clip]{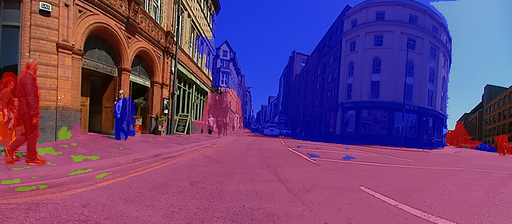}  
    \includegraphics[width=0.245\textwidth, clip]{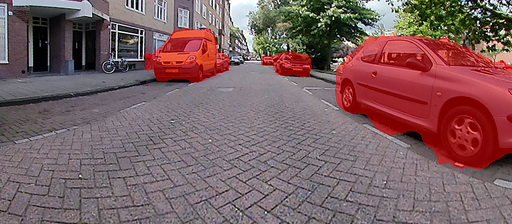}
    \includegraphics[width=0.245\textwidth, clip]{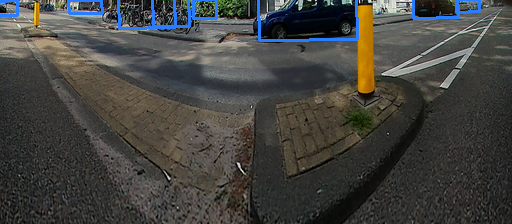}   
    %\rule{\linewidth}{0.5mm}
    
    \vspace{1mm}
    
    % \caption{Whitebox Untargeted Attacks}
%     \label{fig:whitebox_untar}
% \end{figure*}

% \begin{figure*}[t]
%     \centering
    
%     \includegraphics[width=0.245\textwidth,clip]{samples/W213-263.YUV.4CAM.DAT.20180820.132129_MVR_751/cropped.png}
%     \includegraphics[width=0.245\textwidth ,clip]{samples/W222.FR_2014_42a.YUV.4CAM.FR.XCP_USS.LR.DAT.20150714.094534_FV_1291/cropped.png}  
%     \includegraphics[width=0.245\textwidth ,clip]{samples/MAGOTAN-G8A0.PQ35_ACAN.YUV.4CAM.CAN.GPS.DAT.20180621.122200_MVL_481/cropped.png}
%     \includegraphics[width=0.245\textwidth ,clip]{samples/W213-263.YUV.4CAM.DAT.20180523.135257_RV_781/cropped.png}   
    
% \vspace{.5mm}
%   \includegraphics[width=0.245\textwidth ,clip]{samples/W213-263.YUV.4CAM.DAT.20180820.132129_MVR_751/adv_x_wb_depth_tar.png}
%     \includegraphics[width=0.245\textwidth ,clip]{samples/W222.FR_2014_42a.YUV.4CAM.FR.XCP_USS.LR.DAT.20150714.094534_FV_1291/adv_x_wb_ss_tar.png}  
%     \includegraphics[width=0.245\textwidth ,clip]{samples/MAGOTAN-G8A0.PQ35_ACAN.YUV.4CAM.CAN.GPS.DAT.20180621.122200_MVL_481/adv_x_wb_motion_tar.png}
%     \includegraphics[width=0.245\textwidth ,clip]{samples/W213-263.YUV.4CAM.DAT.20180523.135257_RV_781/adv_x_wb_od_tar.png} 
    
%\vspace{.5mm}
    \includegraphics[width=0.245\textwidth ,clip]{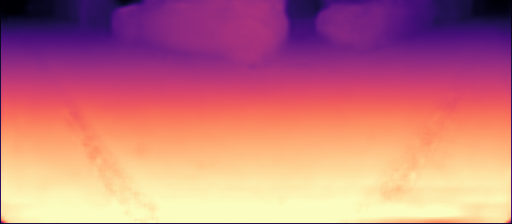}
    \includegraphics[width=0.245\textwidth, clip]{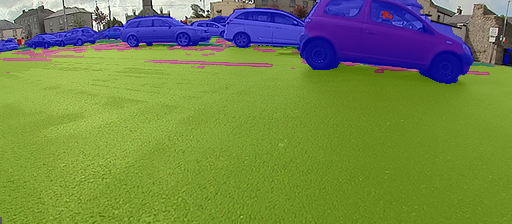}  
    \includegraphics[width=0.245\textwidth, clip]{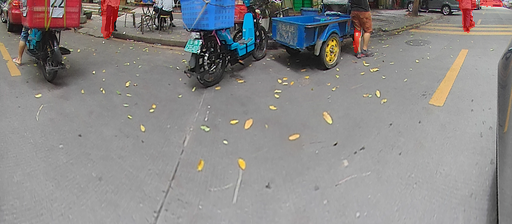}
    \includegraphics[width=0.245\textwidth, clip]{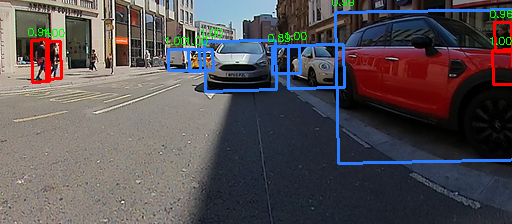}  
    
% \vspace{.5mm}
    \includegraphics[width=0.245\textwidth ,clip]{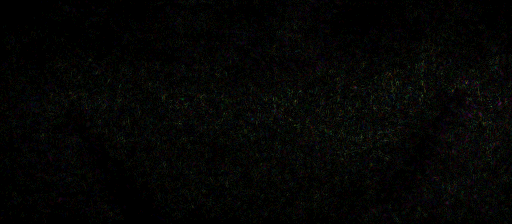}
    \includegraphics[width=0.245\textwidth ,clip]{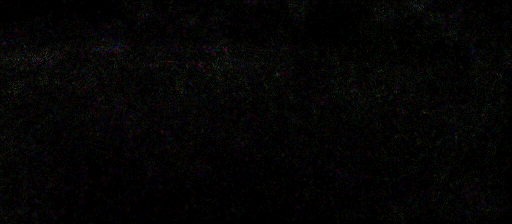}  
    \includegraphics[width=0.245\textwidth ,clip]{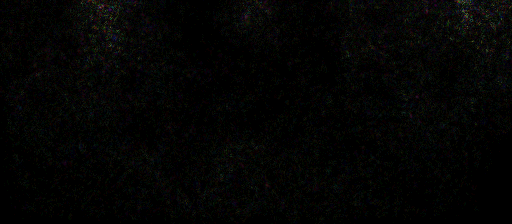}
    \includegraphics[width=0.245\textwidth,clip]{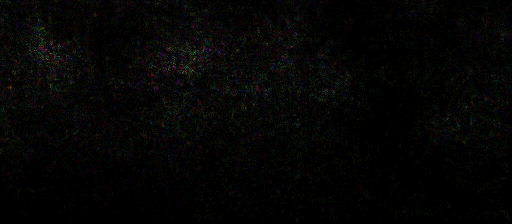}   
    
%\vspace{.5mm}
    \includegraphics[width=0.245\textwidth, clip]{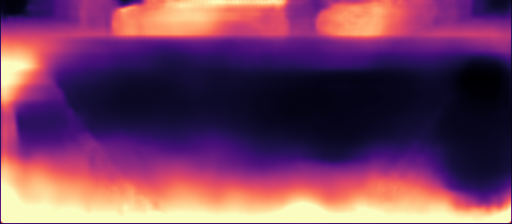}
    \includegraphics[width=0.245\textwidth,clip]{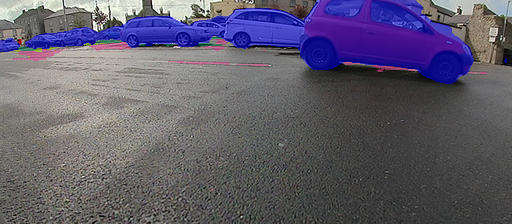}  
    \includegraphics[width=0.245\textwidth, clip]{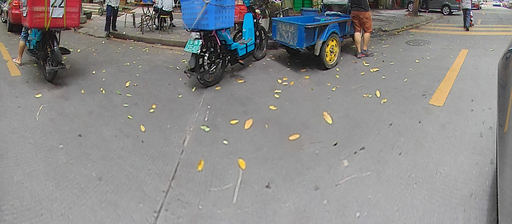}
    \includegraphics[width=0.245\textwidth, clip]{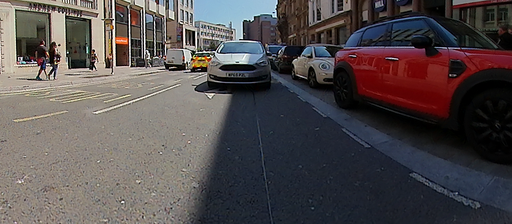}  
    % \vspace{0.5pt}    
    
    % \caption{Whitebox Targeted Attacks}
%     \label{fig:whitebox_tar}
% \end{figure*}

    \vspace{1mm}

% \begin{figure*}[t]
%     \centering
    
%     \includegraphics[width=0.245\textwidth,clip]{samples/W213-263.YUV.4CAM.DAT.20180822.114732_RV_271/cropped.png}
%     \includegraphics[width=0.245\textwidth ,clip]{samples/W213-263.YUV.4CAM.DAT.20180601.170132_MVL_691/cropped.png}  
%     \includegraphics[width=0.245\textwidth ,clip]{samples/W213-263.YUV.4CAM.DAT.20180523.130721_FV_601/cropped.png}
%     \includegraphics[width=0.245\textwidth ,clip]{samples/W222.K_Matrix_42a.YUV.4CAM.FR.XCP_USS.LR.DAT.20150227.142156_MVR_271/cropped.png}   
    
% \vspace{.5mm}
%   \includegraphics[width=0.245\textwidth ,clip]{samples/W213-263.YUV.4CAM.DAT.20180822.114732_RV_271/adv_x_bb_depth_untar.png}
%     \includegraphics[width=0.245\textwidth ,clip]{samples/W213-263.YUV.4CAM.DAT.20180601.170132_MVL_691/adv_x_bb_ss_untar.png}  
%     \includegraphics[width=0.245\textwidth ,clip]{samples/W213-263.YUV.4CAM.DAT.20180523.130721_FV_601/adv_x_bb_motion_untar.png}
%     \includegraphics[width=0.245\textwidth ,clip]{samples/W222.K_Matrix_42a.YUV.4CAM.FR.XCP_USS.LR.DAT.20150227.142156_MVR_271/adv_x_bb_od_untar.png} 

%\vspace{.5mm}
    \includegraphics[width=0.245\textwidth ,clip]{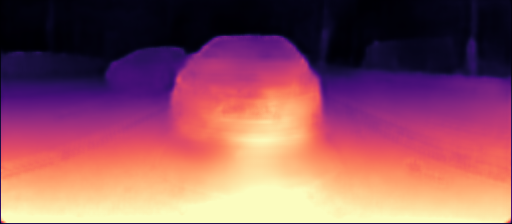}
    \includegraphics[width=0.245\textwidth, clip]{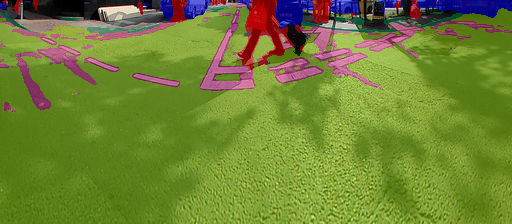}  
    \includegraphics[width=0.245\textwidth, clip]{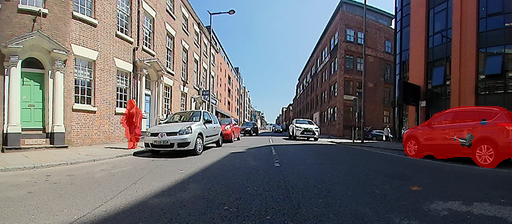}
    \includegraphics[width=0.245\textwidth, clip]{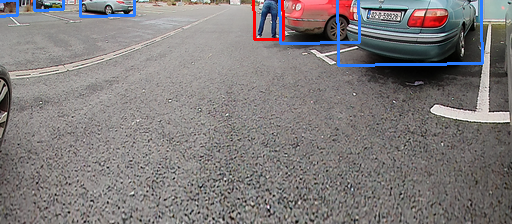}  
    
% \vspace{.5mm}
    \includegraphics[width=0.245\textwidth ,clip]{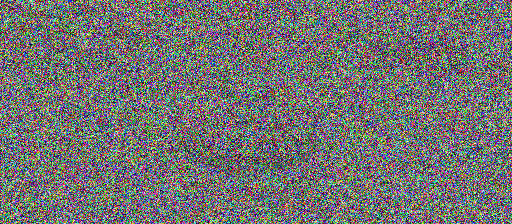}
    \includegraphics[width=0.245\textwidth ,clip]{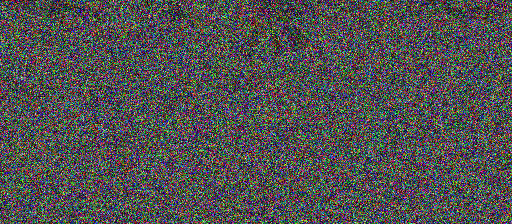}  
    \includegraphics[width=0.245\textwidth ,clip]{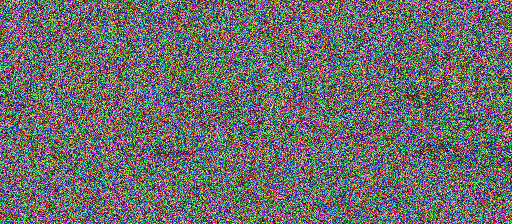}
    \includegraphics[width=0.245\textwidth, clip]{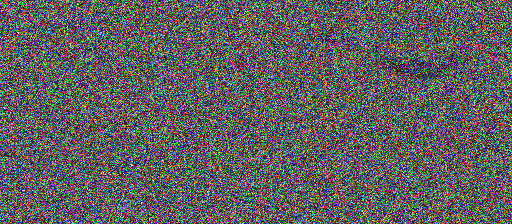}

% \vspace{.5mm}
    \includegraphics[width=0.245\textwidth, clip]{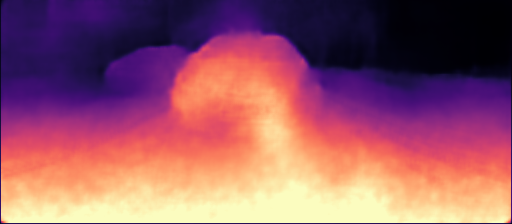}
    \includegraphics[width=0.245\textwidth,clip]{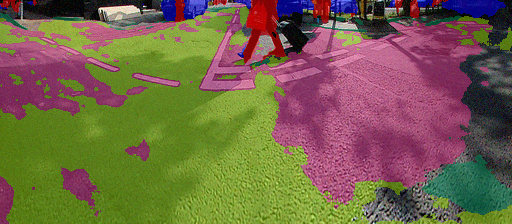}  
    \includegraphics[width=0.245\textwidth, clip]{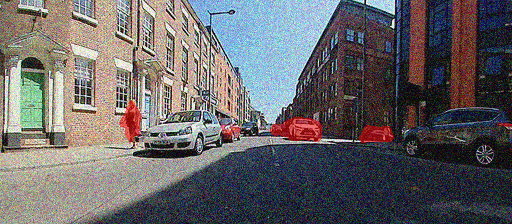}
    \includegraphics[width=0.245\textwidth, clip]{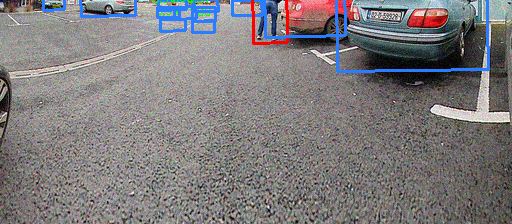}   
    % \vspace{0.5pt}    
    
    % \caption{Blackbox Untargeted Attacks}
%     \label{fig:blackbox_untar}
% \end{figure*}
    
    \vspace{1mm}

% \begin{figure*}[t]
%     \centering
    
%     \includegraphics[width=0.245\textwidth,clip]{samples/W222.K_Matrix_42a.YUV.4CAM.FR.XCP_USS.LR.DAT.20150224.125806_MVL_1351/cropped.png}
%     \includegraphics[width=0.245\textwidth ,clip]{samples/W213-263.YUV.4CAM.DAT.20180821.104931_RV_571/cropped.png}  
%     \includegraphics[width=0.245\textwidth ,clip]{samples/W213-263.YUV.4CAM.DAT.20180529.135903_MVR_241/cropped.png}
%     \includegraphics[width=0.245\textwidth ,clip]{samples/MAGOTAN-G8A0.PQ35_ACAN.YUV.4CAM.CAN.GPS.DAT.20180614.155638_FV_601/cropped.png}   
    
% \vspace{.5mm}
%   \includegraphics[width=0.245\textwidth ,clip]{samples/W222.K_Matrix_42a.YUV.4CAM.FR.XCP_USS.LR.DAT.20150224.125806_MVL_1351/adv_x_bb_depth_tar.png}
%     \includegraphics[width=0.245\textwidth ,clip]{samples/W213-263.YUV.4CAM.DAT.20180821.104931_RV_571/adv_x_bb_ss_tar.png}  
%     \includegraphics[width=0.245\textwidth ,clip]{samples/W213-263.YUV.4CAM.DAT.20180529.135903_MVR_241/adv_x_bb_motion_tar.png}
%     \includegraphics[width=0.245\textwidth ,clip]{samples/MAGOTAN-G8A0.PQ35_ACAN.YUV.4CAM.CAN.GPS.DAT.20180614.155638_FV_601/adv_x_bb_od_tar.png} 

%\vspace{.5mm}
    \includegraphics[width=0.245\textwidth ,clip]{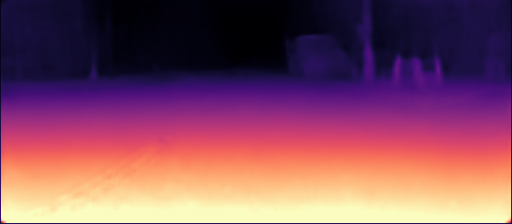}
    \includegraphics[width=0.245\textwidth, clip]{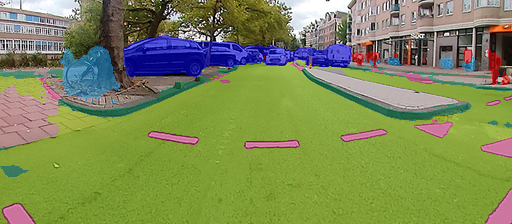}  
    \includegraphics[width=0.245\textwidth, clip]{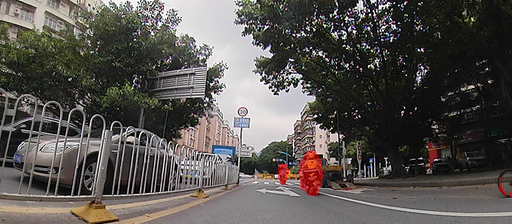}
    \includegraphics[width=0.245\textwidth, clip]{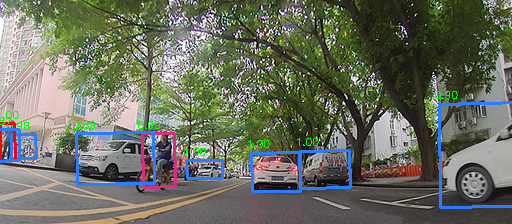}  
    
% \vspace{.5mm}
    \includegraphics[width=0.245\textwidth ,clip]{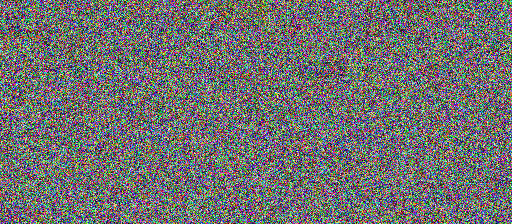}
    \includegraphics[width=0.245\textwidth ,clip]{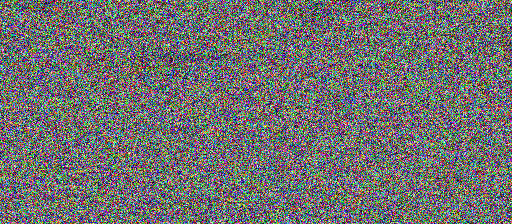}  
    \includegraphics[width=0.245\textwidth ,clip]{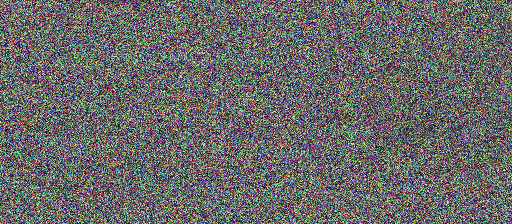}
    \includegraphics[width=0.245\textwidth,clip]{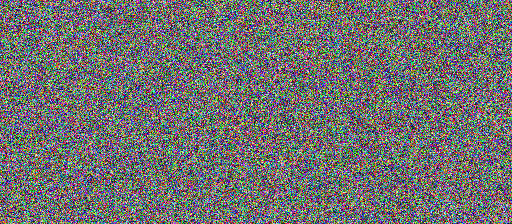}

%\vspace{.5mm}
    \includegraphics[width=0.245\textwidth, clip]{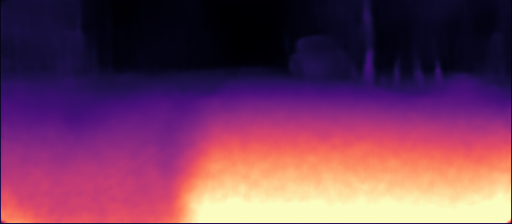}
    \includegraphics[width=0.245\textwidth,clip]{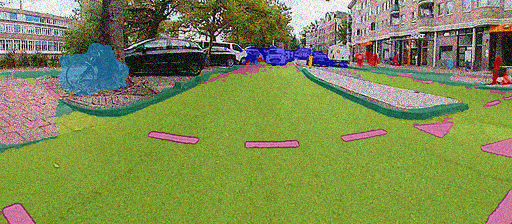}  
    \includegraphics[width=0.245\textwidth, clip]{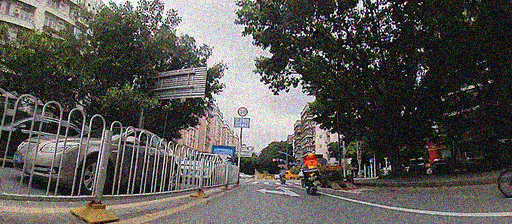}
    \includegraphics[width=0.245\textwidth, clip]{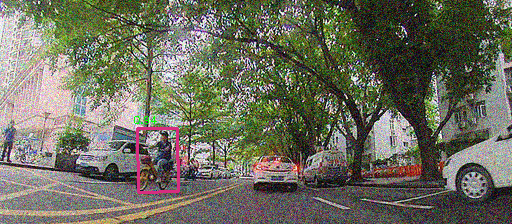}   
    
    \caption{From Top to Bottom: \textbf{White box Un-targeted, White box Targeted, Black box Un-targeted, \& Black box Targeted Attacks.} Within each group from top to bottom: Original results, adversarial perturbations, \& the impacted results.}
    \label{fig:samples}
\end{figure*}

% -------------------------------------------------
\begin{table}[t]
\captionsetup{singlelinecheck=false, skip=0pt, font=small, belowskip=0pt}
\centering
\caption{\textbf{Summary of attacking and defending results} across the test data where \textit{A} and \textit{D} columns are for Attack and Defense respectively.}
\begin{adjustbox}{width=\columnwidth}
\small
\setlength{\tabcolsep}{0.3em}
\begin{tabular}{l|l|ll|ll|ll|ll}
\toprule
\textbf{Task} &  
& \multicolumn{2}{c}{\textit{\begin{tabular}[c]{@{}c@{}} \cellcolor[HTML]{00b0f0} Distance\\ \cellcolor[HTML]{00b0f0} RMSE\end{tabular}}} 
& \multicolumn{2}{c}{\textit{\begin{tabular}[c]{@{}c@{}} \cellcolor[HTML]{00b050} Segmentation\\ \cellcolor[HTML]{00b050} mIoU\end{tabular}}} 
& \multicolumn{2}{c}{\textit{\begin{tabular}[c]{@{}c@{}}
\cellcolor[HTML]{ab9ac0} Motion\\ \cellcolor[HTML]{ab9ac0} mIoU\end{tabular}}} 
& \multicolumn{2}{c}{\textit{\begin{tabular}[c]{@{}c@{}}
\cellcolor[HTML]{a5a5a5} Detection\\ \cellcolor[HTML]{a5a5a5} mAP\end{tabular}}} \\ 
\midrule
& \multicolumn{1}{l|}{} & \multicolumn{1}{c|}{\cellcolor[HTML]{7d9ebf} A} & \multicolumn{1}{c|}{\cellcolor[HTML]{e8715b} D} 
& \multicolumn{1}{l|}{\cellcolor[HTML]{7d9ebf} A(\%)} & \multicolumn{1}{l|}{\cellcolor[HTML]{e8715b} D(\%)}
& \multicolumn{1}{l|}{\cellcolor[HTML]{7d9ebf} A(\%)} & \multicolumn{1}{l|}{\cellcolor[HTML]{e8715b} D(\%)} 
& \multicolumn{1}{l|}{\cellcolor[HTML]{7d9ebf} A(\%)} & \multicolumn{1}{l|}{\cellcolor[HTML]{e8715b} D(\%)} \\ 
\cmidrule(l){3-10}
\multirow{4}{*}{\textit{Distance}} 
& \textit{wb\_untarget} & 0.126 & 0.047 & -14 & -7.3 & -3.1 & -3.0 & -13.4 & -25.5 \\
& \textit{wb\_target} & 0.288 & 0.031 & -40 & -7.5 & -4.4 & -2.7 & -38.9 & -33.2 \\
& \textit{bb\_untarget} & 0.036 & 0.033 & -3.0 & -6.5 & -1.5 & -3.7 & -4.6 & -25.8 \\
& \textit{bb\_target} & 0.035 & 0.036 & -14.8 & -13.4 & -3.9 & -2.9 & -27.1 & -37.6 \\
\midrule
\multirow{4}{*}{\textit{Segmentation}} 
& \textit{wb\_untarget} & 0.032 & 0.028 & -86.8 & -14.2 & -5.0 & -3.5 & -37.6 & -30.1 \\
& \textit{wb\_target} & 0.017 & 0.027 & -32.0 & -5.5 & -4.0 & -2.6 & -21.3 & -27.5 \\
& \textit{bb\_untarget} & 0.015 & 0.031 & -26.1 & -11.4 & -2.3 & -4.9 & -6.7 & -27.3 \\
& \textit{bb\_target} & 0.020 & 0.034 & -16.1 & -9.2 & -2.2 & -2.5 & 9.2 &  -28.8 \\
\midrule
\multirow{4}{*}{\textit{Motion}} 
& \textit{wb\_untarget} & 0.018 & 0.027 & -11.1 & -7.3 & -25.9 & -9.2 & -18.9 & -23.1 \\
& \textit{wb\_target}   & 0.010 & 0.027 & -2.4 & -6.0 & -14.7 & -9.2 & -7.1 & -30.0 \\
& \textit{bb\_untarget} & 0.030 & 0.039 & -17.1 & -15.8 & -24.3 & -17.6 & -22.2 & -38.6 \\
& \textit{bb\_target}   & 0.033 & 0.040 & -24.6 & -22.5 & -13.9 & -11.7 & -34.7 & -47.9 \\
\midrule
\multirow{4}{*}{\textit{Detection}} 
& \textit{wb\_untarget} & 0.012 & 0.027 & -5.1 & -5.9 & -2.1 & -2.5 & -39.8 & -31.4 \\
& \textit{wb\_target} & 0.018 & 0.027 & -15.0 & -6.2 & -3.0 & -4.0 & -71.9 & -30.6 \\
& \textit{bb\_untarget} & 0.021 & 0.033 & -12.5 & -10.4 & -4.2 & -5.8 & -39.4 & -35.3 \\
& \textit{bb\_target} & 0.022 & 0.034 & -11.6 & -10.6 & -2.7 &
-3.7 & -34.4 & -37.9 \\ 
\bottomrule
\end{tabular}
\end{adjustbox}
\label{tab:table-attackdefense1}
\end{table}
% -------------------------------------------------
To understand the effect of applying a defense method on the attacks, Gaussian blurring with radius = $1$ is applied to the final adversarial examples and then fed into the target network, and performance is reported. As shown in Table \ref{tab:table-attackdefense1}, this simple defense method has a positive effect for both segmentation and motion tasks in most cases compared to depth and detection tasks. Furthermore, the effect of blurring on the network's performance is inspected without applying any attacks, as shown in Table \ref{tab:table-blur}. Both detection and distance tasks are affected the most. This explains why this defense method is more effective for segmentation and motion tasks. 
Figure \ref{fig:samples} shows different visual samples of the attacks organized into four groups. Each group has three images: the original output, the adversarial perturbations magnified to 10X, and the impacted results are overplayed on the adversarial examples. As expected, perturbations for the white box attacks are much more minor and more localized than the black box case. Moreover, for the un-targeted attacks, the performance is harmed without having a specific goal leading to arbitrary predictions. On the other hand, for targeted attacks, vehicles or roads are removed for the semantic segmentation task. We add false objects or remove true objects for the detection task. Near pixels are converted as far for the distance task. Finally, we convert dynamic objects to static for the motion task.\par
% -------------------------------------------------

% -------------------------------------------------
\begin{table}[t]
\captionsetup{singlelinecheck=false, font=small, belowskip=-10pt}
\caption{\textbf{Input blurring effect on the tasks.}}
\begin{center}
\begin{tabular}{@{}l|l|c|c|c@{}}
\toprule
\textbf{Task } & \textit{\cellcolor[HTML]{7d9ebf} Metric} & \multicolumn{1}{l|}{\textit{\cellcolor[HTML]{7d9ebf} Original}} & \multicolumn{1}{l|}{\textit{\cellcolor[HTML]{e8715b} Blurred}} & \multicolumn{1}{l}{\textit{\cellcolor[HTML]{e8715b} Effect (\%)}} \\ \midrule
\textit{Distance} & RMSE & 0.0 & 0.026 & NA \\ 
\textit{Segmentation} & mIoU & 0.499 & 0.477 &  -4.4 \\ 
\textit{Motion} & mIoU  & 0.711 & 0.693 & -2.6 \\ 
\textit{Detection} & mAP & 0.633 & 0.416 & -34.3 \\ 
\bottomrule
\end{tabular}
\label{tab:table-blur}
\end{center}
\end{table}
% -------------------------------------------------
% -------------------------------------------------
% -------------------------------------------------
\section{Conclusion}

In this work, various adversarial attacks are applied on a multi-task target network with shared encoder and different decoders for autonomous driving visual perception. For each perception task, white and black box attacks are conducted for targeted and un-targeted scenarios. Moreover, attacking curves show the interactions between the attacks on different tasks. It is shown how attacking a task has an effect not only on that task but also on the others. Moreover, by applying blurring on the adversarial examples as a defense method, it is found to have a positive effect on segmentation and motion tasks in contrast to object detection and distance tasks for the considered target network. In the future, we plan to conduct physical attacks, try other sensors such as LiDAR, and attacking multiple tasks jointly. It is obvious that, attacks and defenses are still challenging tasks and an active area of research, especially for autonomous driving applications with multi-task deep networks.\par
% -------------------------------------------------
\begin{biography}

\vspace{2pt}

\textbf{{Ibrahim Sodh}} Ibrahim has 23 years of experience in the area of Machine Learning and Software Development. He received his PhD degree in Deep Reinforcement Learning and computer vision. His M.Sc. Thesis is in Machine Learning applied for automatic documents summarization. Ibrahim has participated in several related national and international mega projects, conferences and summits. He delivers training and lectures for academic and industrial entities. His patents and publications are mainly in Natural language processing, Speech processing, Computer vision. Currently Ibrahim Sobh is a Senior Expert of AI, Valeo Group." \par

\textbf{{Ahmed Hamed}} received a B.E. degree in 2019, He joined Valeo Egypt in November 2019 as a Machine Learning Engineer. He worked on both research and industrial projects, His research focuses mainly on deep learning and computer vision, he has also received Udacity's Self-driving Car Nano Degree in 2020 and currently he is studying for M.Sc. degree.

\textbf{Varun Ravi Kumar} received a B.E.\ degree in 2015 and an M.Sc.\ degree in 2017 from TU Chemnitz, Germany. He is currently a Ph.D. student in Deep Learning for autonomous driving affiliated to TU Ilmenau and is currently working at Valeo. His research is mainly focused on the design of self-supervised perception algorithms using neural networks for self-driving cars. His expertise lies in depth and flow estimation for fisheye images and multi-task modeling. His focus also lies in semantic, motion segmentation, 2D and 3D object detection, and point cloud processing. He was awarded the \emph{Deutschlandstipendium} for top-class international talent. He was also part of Udacity's first cohort of Self-Driving-Car Nanodegree in 2017.\par

\textbf{Senthil Yogamani} is an Artificial Intelligence architect and holds a director level technical leader position at Valeo Ireland. He leads the research and design of AI algorithms for various modules of autonomous driving systems. He has over 15 years of experience in computer vision and machine learning including 13 years of experience in industrial automotive systems. He is an author of 100+ publications with 2300+ citations and 100+ inventions with 70 filed patent families. He serves on the editorial board of various leading IEEE automotive conferences including ITSC and IV and advisory board of various industry consortia including Khronos, Cognitive Vehicles and IS Auto. He is a recipient of the best associate editor award at ITSC 2015 and best paper award at ITST 2012.\par

\end{biography}
% -------------------------------------------------
{\small
\bibliographystyle{bib/ieee_fullname}
\bibliography{bib/egbib}

\begin{thebibliography}{10}\itemsep=-1pt

\bibitem{athalye2018synthesizing}
Anish Athalye, Logan Engstrom, Andrew Ilyas, and Kevin Kwok.
\newblock {Synthesizing robust adversarial examples}.
\newblock In {\em Proc. of ICML}, pages 284--293. PMLR, 2018.

\bibitem{berman2018lovasz}
Maxim Berman, Amal~Rannen Triki, and Matthew~B Blaschko.
\newblock {{The lov{\'a}sz-softmax loss: A tractable surrogate for the
  optimization of the intersection-over-union measure in neural networks}}.
\newblock In {\em Proc. of CVPR}, 2018.

\bibitem{cao2019adversarial}
Yulong Cao, Chaowei Xiao, Dawei Yang, Jing Fang, Ruigang Yang, et~al.
\newblock {Adversarial objects against lidar-based autonomous driving systems}.
\newblock {\em arXiv preprint arXiv:1907.05418}, 2019.

\bibitem{Chennupati_2019}
S. Chennupati, Ganesh Sistu., Senthil Yogamani., and Samir Rawashdeh.
\newblock Auxnet: Auxiliary tasks enhanced semantic segmentation for automated
  driving.
\newblock In {\em Proceedings of the 14th International Joint Conference on
  Computer Vision, Imaging and Computer Graphics Theory and Applications:
  VISAPP}, 2019.

\bibitem{dahal2021roadedgenet}
Ashok Dahal, Eric Golab, Rajender Garlapati, Varun Ravi~Kumar, and Senthil
  Yogamani.
\newblock {RoadEdgeNet: Road Edge Detection System Using Surround View Camera
  Images}.
\newblock In {\em Electronic Imaging}. Society for Imaging Science and
  Technology, 2021.

\bibitem{dahal2021online}
Ashok Dahal, Varun~Ravi Kumar, Senthil Yogamani, and Ciaran Eising.
\newblock An online learning system for wireless charging alignment using
  surround-view fisheye cameras.
\newblock {\em arXiv preprint arXiv:2105.12763}, 2021.

\bibitem{das2020tiledsoilingnet}
Arindam Das, Pavel K{\v{r}}{\'\i}{\v{z}}ek, Ganesh Sistu, Fabian B{\"u}rger,
  et~al.
\newblock {TiledSoilingNet: Tile-level Soiling Detection on Automotive
  Surround-view Cameras Using Coverage Metric}.
\newblock In {\em Proc. of ITSC}, pages 1--6. IEEE, 2020.

\bibitem{dhananjaya2021weather}
Mahesh~M Dhananjaya, Varun~Ravi Kumar, and Senthil Yogamani.
\newblock {Weather and Light Level Classification for Autonomous Driving:
  Dataset, Baseline and Active Learning}.
\newblock In {\em Proc. of ITSC}. IEEE, 2021.

\bibitem{dziugaite2016study}
Gintare~Karolina Dziugaite, Zoubin Ghahramani, and Daniel~M Roy.
\newblock {A study of the effect of jpg compression on adversarial images}.
\newblock {\em arXiv preprint arXiv:1608.00853}, 2016.

\bibitem{eykholt2018robust}
Kevin Eykholt, Ivan Evtimov, Earlence Fernandes, Bo Li, Amir Rahmati, Chaowei
  Xiao, Atul Prakash, Tadayoshi Kohno, and Dawn Song.
\newblock {Robust physical-world attacks on deep learning visual
  classification}.
\newblock In {\em Proc. of CVPR}, 2018.

\bibitem{gallagher2021hybrid}
Louis Gallagher, Varun~Ravi Kumar, Senthil Yogamani, and John~B McDonald.
\newblock A hybrid sparse-dense monocular slam system for autonomous driving.
\newblock In {\em Proc. of ECMR}, pages 1--8. IEEE, 2021.

\bibitem{goodfellow2014generative}
Ian~J Goodfellow, Jean Pouget-Abadie, Mehdi Mirza, Bing Xu, David Warde-Farley,
  Sherjil Ozair, Aaron Courville, and Yoshua Bengio.
\newblock {Generative adversarial networks}.
\newblock {\em arXiv preprint arXiv:1406.2661}, 2014.

\bibitem{goodfellow2015explaining}
Ian~J Goodfellow, Jonathon Shlens, et~al.
\newblock {Explaining and harnessing adversarial examples}.
\newblock In {\em Proc. of ICLR}, 2015.

\bibitem{kia_2021}
Sebastian Houben, Stephanie Abrecht, Maram Akila, Andreas B{\"{a}}r, et~al.
\newblock {Inspect, Understand, Overcome: A Survey of Practical Methods for AI
  Safety}.
\newblock {\em CoRR}, abs/2104.14235, 2021.

\bibitem{kumar2020fisheyedistancenet}
Varun~Ravi Kumar, Sandesh~Athni Hiremath, Markus Bach, Stefan Milz, et~al.
\newblock {Fisheyedistancenet: Self-supervised scale-aware distance estimation
  using monocular fisheye camera for autonomous driving}.
\newblock In {\em Proc. of ICRA}. IEEE, 2020.

\bibitem{kumar2021syndistnet}
Varun~Ravi Kumar, Marvin Klingner, Senthil Yogamani, Stefan Milz, Tim
  Fingscheidt, and Patrick Mader.
\newblock {SynDistNet: Self-supervised monocular fisheye camera distance
  estimation synergized with semantic segmentation for autonomous driving}.
\newblock In {\em Proc. of WACV}, pages 61--71, 2021.

\bibitem{kumar2018monocular}
Varun~Ravi Kumar, Stefan Milz, Christian Witt, Martin Simon, Karl Amende,
  Johannes Petzold, Senthil Yogamani, and Timo Pech.
\newblock Monocular fisheye camera depth estimation using sparse lidar
  supervision.
\newblock In {\em Proc. of ITSC}. IEEE, 2018.

\bibitem{kumar2021omnidet}
Varun~Ravi Kumar, Senthil Yogamani, Hazem Rashed, Ganesh Sitsu, et~al.
\newblock Omnidet: Surround view cameras based multi-task visual perception
  network for autonomous driving.
\newblock {\em Proc. of ICRA + RA-L}, 6(2):2830--2837, 2021.

\bibitem{leang2020dynamic}
Isabelle Leang, Ganesh Sistu, Fabian B{\"u}rger, Andrei Bursuc, and Senthil
  Yogamani.
\newblock Dynamic task weighting methods for multi-task networks in autonomous
  driving systems.
\newblock In {\em Proc. of ITSC}, pages 1--8. IEEE, 2020.

\bibitem{lin2017focal}
Tsung-Yi Lin, Priya Goyal, Ross Girshick, Kaiming He, and Piotr Doll{\'a}r.
\newblock {Focal loss for dense object detection}.
\newblock In {\em Proc. of CVPR}, pages 2980--2988, 2017.

\bibitem{mao2020multitask}
Chengzhi Mao, Amogh Gupta, Vikram Nitin, Baishakhi Ray, Shuran Song, Junfeng
  Yang, and Carl Vondrick.
\newblock {Multitask Learning Strengthens Adversarial Robustness}.
\newblock In {\em ECCV 2020}, pages 158--174. Springer International
  Publishing, 2020.

\bibitem{rashed2021generalized}
Hazem Rashed, Eslam Mohamed, Ganesh Sistu, Varun~Ravi Kumar, Ciaran Eising,
  Ahmad El-Sallab, and Senthil Yogamani.
\newblock {Generalized object detection on fisheye cameras for autonomous
  driving: Dataset, representations and baseline}.
\newblock In {\em Proceedings of the IEEE/CVF Winter Conference on Applications
  of Computer Vision}, pages 2272--2280, 2021.

\bibitem{rashedfisheyeyolo}
Hazem Rashed, Eslam Mohamed, Varun Ravi~Kumar Sistu, Ganesh~and, Ciar{\'a}n
  Eising, Ahmad El-Sallab, and Senthil Yogamani.
\newblock {FisheyeYOLO: Object Detection on Fisheye Cameras for Autonomous
  Driving}.
\newblock {\em Machine Learning for Autonomous Driving NeurIPS 2020 Virtual
  Workshop}, 2020.

\bibitem{kumar2021svdistnet}
Varun Ravi~Kumar, Marvin Klingner, Senthil Yogamani, Markus Bach, Stefan Milz,
  Tim Fingscheidt, and Patrick M\"{a}der.
\newblock {SVDistNet: Self-Supervised Near-Field Distance Estimation on
  Surround View Fisheye Cameras}.
\newblock {\em Proc. of T-ITS}, abs/2104.04420, 2021.

\bibitem{kumar2018near}
Varun Ravi~Kumar, Stefan Milz, Christian Witt, Martin Simon, Karl Amende,
  Johannes Petzold, Senthil Yogamani, and Timo Pech.
\newblock {Near-field depth estimation using monocular fisheye camera: A
  semi-supervised learning approach using sparse LiDAR data}.
\newblock In {\em CVPR Workshop}, volume~7, 2018.

\bibitem{kumar2020unrectdepthnet}
Varun Ravi~Kumar, Senthil Yogamani, Markus Bach, Christian Witt, Stefan Milz,
  and Patrick M{\"{a}}der.
\newblock {UnRectDepthNet: Self-Supervised Monocular Depth Estimation using a
  Generic Framework for Handling Common Camera Distortion Models}.
\newblock In {\em Proc. of IROS}, pages 8177--8183, 2020.

\bibitem{kumar2021fisheyedistancenet++}
Varun Ravi~Kumar, Senthil Yogamani, Stefan Milz, and Patrick M\"{a}der.
\newblock {FisheyeDistanceNet++: Self-Supervised Fisheye Distance Estimation
  with Self-Attention, Robust Loss Function and Camera View Generalization}.
\newblock In {\em Electronic Imaging}. Society for Imaging Science and
  Technology, 2021.

\bibitem{samangouei2018defense}
Pouya Samangouei, Maya Kabkab, and Rama Chellappa.
\newblock {Defense-GAN: Protecting Classifiers Against Adversarial Attacks
  Using Generative Models}.
\newblock In {\em Proc. of ICLR}, 2018.

\bibitem{shu2020feature}
Chang Shu, Kun Yu, Zhixiang Duan, and Kuiyuan Yang.
\newblock {Feature-metric loss for self-supervised learning of depth and
  egomotion}.
\newblock In {\em Proc. of ECCV}, pages 572--588. Springer, 2020.

\bibitem{sistu2019real}
Ganesh Sistu, Isabelle Leang, and Senthil Yogamani.
\newblock Real-time joint object detection and semantic segmentation network
  for automated driving.
\newblock {\em Proceedings of NeurIPS workshop on Machine Learning for
  Autonomous Driving}, 2019.

\bibitem{thys2019fooling}
Simen Thys, Wiebe Van~Ranst, and Toon Goedem{\'e}.
\newblock {Fooling automated surveillance cameras: adversarial patches to
  attack person detection}.
\newblock In {\em Proc. of CVPR Workshops}, pages 49--55, 2019.

\bibitem{uricar2021let}
Michal Uricar, Ganesh Sistu, Hazem Rashed, Varun Ravi~Kumar Vobecky,
  Antonin~and, Pavel Krizek, Fabian Burger, and Senthil Yogamani.
\newblock {Let's Get Dirty: GAN Based Data Augmentation for Camera Lens Soiling
  Detection in Autonomous Driving}.
\newblock In {\em Proc. of WACV}, pages 766--775, 2021.

\bibitem{avsafetywho}
{WHO}.
\newblock {Global status report on road safety 2018, ''World Health
  Organization}.
\newblock
  \url{https://apps.who.int/iris/bitstream/handle/10665/276462/9789241565684-eng.pdf}.
\newblock [Accessed October-2020].

\bibitem{xie2017adversarial}
Cihang Xie, Jianyu Wang, Zhishuai Zhang, Yuyin Zhou, Lingxi Xie, and Alan
  Yuille.
\newblock {Adversarial examples for semantic segmentation and object
  detection}.
\newblock In {\em Proc. of ICCV}, pages 1369--1378, 2017.

\bibitem{xu2018feature}
Weilin Xu, David Evans, and Yanjun Qi.
\newblock {Feature squeezing: Detecting adversarial examples in deep neural
  networks}.
\newblock In {\em 25th Annual Network and Distributed System Security
  Symposium}. The Internet Society, 2018.

\bibitem{yahiaoui2019fisheyemodnet}
Marie Yahiaoui, Hazem Rashed, Letizia Mariotti, Ganesh Sistu, Ian Clancy, Lucie
  Yahiaoui, Varun Ravi~Kumar, and Senthil Yogamani.
\newblock {FisheyeModNet: Moving object detection on Surround-View Cameras for
  Autonomous Driving}.
\newblock {\em arXiv preprint arXiv:1908.11789}, 2019.

\bibitem{yogamani2019woodscape}
Senthil Yogamani, Ciar{\'a}n Hughes, Jonathan Horgan, Ganesh Sistu, Padraig
  Varley, Derek O'Dea, et~al.
\newblock {Woodscape: A multi-task, multi-camera fisheye dataset for autonomous
  driving}.
\newblock In {\em Proc. of ICCV}, pages 9308--9318, 2019.

\end{thebibliography}
}
\end{document}